\documentclass[pdflatex,sn-mathphys-num]{sn-jnl}% Math and Physical Sciences Numbered Reference Style
\usepackage[T1]{fontenc}
\DeclareFontFamily{U}{rsfs}{}
\DeclareFontShape{U}{rsfs}{m}{n}{<-> s * [1.2] rsfs10}{}
\usepackage{graphicx}%
\usepackage{multirow}%
\usepackage{amsmath,amssymb,amsfonts}%
\usepackage{amsthm}%
\usepackage{float}
\usepackage{mathrsfs}%
\usepackage[title]{appendix}%
\usepackage{xcolor}%
\usepackage{textcomp}%
\usepackage{manyfoot}%
\usepackage{booktabs}
\usepackage{multirow}
\usepackage{multicol, makecell}
\usepackage{tabularx}
\usepackage{longtable}
\usepackage{booktabs}%
\usepackage{algorithm}%
\usepackage{algorithmicx}%
\usepackage{algpseudocode}%
\usepackage{listings}%
\usepackage{setspace}
\usepackage{xcolor}
\newcommand{\best}[1]{\textbf{#1}}
\newcommand{\second}[1]{\textcolor{blue}{#1}}

\theoremstyle{thmstyleone}%
\begin{document}

\title[A Multimodal Explainable Deep Learning Framework for Alzheimer’s Disease Diagnosis using 3D Magnetic Resonance Imaging and Clinical Data]{Multimodal Cross-Attention  Deep Learning for Explainable Alzheimer's Disease Diagnosis using 3D Magnetic Resonance Imaging and Clinical Data\
}


\author*[1]{\fnm{Yusuf} \sur{Brima}}\email{ybrima@uos.de}

\author*[2,3]{\fnm{Marcellin} \sur{Atemkeng}}
\email{m.atemkeng@ru.ac.za}

\author[4]{\fnm{Lakshmana Rao} \sur{Namamula}}
\author[5]{\fnm{Antoine} \sur{Vacavant}}

\affil*[1]{\orgname{Computer Vision, Institute of Cognitive Science, Osnabr\"{u}ck University}, \orgaddress{\country{Germany}}}

\affil*[2]{\orgname{Department of Mathematics, Rhodes University}, \orgaddress{\country{South Africa}}}

\affil*[3]{\orgname{National Institute for Theoretical and Computational Sciences (NITheCS)}, \orgaddress{\city{Stellenbosch}, \postcode{7600}, \country{South Africa}}}

\affil*[4]{\orgname{Madanapalle Institute of Technology and Science}, \orgaddress{\city{Madanapalle, Andhra Pradesh}, \country{India}}}

\affil[5]{\orgname{Université Clermont Auvergne, Clermont Auvergne INP, CNRS, Institut Pascal, F-63000}, \orgaddress{\city{Clermont-Ferrand}, \country{France}}}


\abstract{
Dementia is a major and growing global health burden, with Alzheimer's disease (AD) accounting for most cases. Timely and accurate diagnosis is central to managing this burden and increasingly depends on integrating complementary clinical and imaging information. Multimodal deep learning can combine these modalities for AD diagnosis, but how its explanations behave across modalities, fusion strategies, and cohorts remains unclear. We developed an explainable multimodal framework pairing a three-dimensional convolutional neural network (3D CNN) encoder for T1-weighted magnetic resonance imaging (MRI) with a feedforward network for harmonized clinical and demographic data, comparing varied model setups on three-way and pairwise diagnostic tasks using 6,479 internal records from the Alzheimer's Disease Neuroimaging Initiative (ADNI) and 1,703 independent records from the Open Access Series of Imaging Studies-3 (OASIS-3). On ADNI, the tabular-only model achieved the highest three-class area under the receiver operating characteristic curve (AUC–ROC; 0.879) and best discriminated cognitively normal (CN) versus mild cognitive impairment (MCI; 0.903), while cross-attention performed best for MCI versus AD (0.861); CN versus AD was highly discriminative overall. On OASIS-3, the vision-only model performed best (three-class AUC–ROC 0.910); CN versus MCI remained difficult, and no fusion strategy consistently outperformed single modalities across tasks and cohorts. SHapley Additive exPlanations (SHAP) and Integrated Gradients identified the Mini-Mental State Examination (MMSE) as the dominant tabular feature in both cohorts, with global feature rankings agreeing strongly in ADNI ($\rho=0.94$) and OASIS-3 ($\rho=0.96$); class activation map (CAM)-based explanations, however, changed with model configuration and cohort. These findings show that multimodal performance and explanations are task, modality, fusion, and cohort-dependent: a dominant cognitive signal persisted across cohorts, but feature contributions and CAM explanations did not, underscoring the need to evaluate explainability under cohort shift rather than as a stable, intrinsic property.}

\keywords{Multimodal Learning, Deep Learning, Alzheimer's Disease, Explainability, Clinical AI}



\maketitle

\section{Introduction}
\label{sec:intro}

Deep learning, a subfield of Machine Learning, has within little more than a decade become a dominant paradigm for extracting rich structures and statistical regularities from complex, often high-dimensional data. Convolutional neural networks (CNNs) and, more recently, transformer-based architectures now underpin state-of-the-art systems for image recognition, segmentation, and generative modeling~\cite{lecun2015deep}, and this representation learning principle has propagated across the full continuum of healthcare, from medical diagnosis and disease prognosis to drug discovery~\cite{topol2019high,jumper2021highly}. At the same time, increasing life expectancy has expanded the population at risk of neuro-cognitive disorders. Age remains the strongest known risk factor for dementia, with prevalence rising sharply beyond 65 years~\cite{grueso2021machine}. Globally, nearly ten million new cases of dementia are diagnosed each year, with an economic burden estimated at US\$1.3 trillion in 2019~\cite{who2025dementia}. Alzheimer's disease (AD) accounts for an estimated 60--70\% of dementia cases, while over 60\% of the 57 million people living with dementia in 2021 resided in low- and middle-income countries (LMICs), where specialist workforces and imaging infrastructure are often constrained~\cite{who2025dementia}. This disparity highlights the need for scalable clinical decision support systems based on routinely collected data.

AD and its prodromal stage, mild cognitive impairment (MCI), are therefore important targets for AI-enabled healthcare, with prevalence projected to reach 139 million by 2050~\cite{world2021global}. Conventional diagnosis often involves extended neuropsychological testing, expert review, and costly or invasive biomarkers such as cerebrospinal fluid assays or amyloid-PET. Structural Magnetic Resonance Imaging (sMRI), together with routinely available clinical and demographic variables, offers a potentially more scalable source of diagnostic information. Early and accurate differentiation between cognitively normal individuals, MCI, and AD is particularly important for timely intervention, yet remains challenging because of the subtle and overlapping symptomatology of MCI. CNNs can learn disease-related atrophy patterns directly from 2D slices or full 3D volumes, but empirical benchmarking has shown that imaging performance is highly sensitive to data leakage and that, under subject-level splits, three-way CN/MCI/AD discrimination from sMRI remains modest, with MCI a persistent bottleneck~\cite{wen2020convolutional}. Tabular models therefore provide an important benchmark for assessing the added value of imaging, although strong performance on cognitive measures may not generalize across cohorts with different recruitment protocols, assessment procedures, and diagnostic criteria~\cite{grueso2021machine,christodoulou2025artificial}.

To determine whether structural MRI provides complementary information beyond clinical assessments, several studies have proposed multimodal architectures that jointly encode imaging and non-imaging data. Venugopalan et al.~\cite{venugopalan2021multimodal} paired 3D CNNs for MRI with stacked denoising autoencoders for genetic and clinical features, while Golovanevsky et al.~\cite{golovanevsky2022multimodal} used cross-modal attention to model interactions between imaging, genetic, and clinical embeddings. Together with the broader multimodal machine learning literature~\cite{baltruvsaitis2018multimodal}, these studies motivate representation-level fusion strategies. However, increased model capacity also increases opacity, motivating the use of eXplainable AI (XAI). This ``black box'' problem is a recognized barrier to the clinical adoption of AI diagnostic systems~\cite{rudin2019stop,brkan2025article}. Methods such as Grad-CAM~\cite{selvaraju2017grad}, Grad-CAM++~\cite{chattopadhay2018grad}, and Score-CAM~\cite{wang2020score}, amongst others\cite{binder2016layer,kapishnikov2019xrai,kapishnikov2021guided,zhang2025finer}, provide image-level attributions, while SHapley Additive exPlanations (SHAP)~\cite{lundberg2017unified} and Integrated Gradients~\cite{sundararajan2017axiomatic} can if not often used to attribute predictions to tabular features. Agreement across attribution methods can itself provide evidence that explanations capture meaningful rather than method-specific signals~\cite{brima2024saliency,singh2025unsupervised}.

Despite this sheer progress, dual-modality attribution has become common in multimodal AD studies without establishing whether the corresponding explanations remain stable across independent cohorts. Khalid et al.~\cite{khalid2025detection} combine Grad-CAM and SHAP across 2D sMRI slices and clinical symptom data, but train the clinical and imaging classifiers separately, leaving their attribution correspondence to a single underlying decision unclear. Sharshar et al.~\cite{sharshar2025not} combine MRI, radiomics, gene expression, and clinical data through cross-attention and apply Grad-CAM and SHAP to the respective branches. Their OmniBrain model nevertheless showed substantial performance degradation under cross-cohort distribution analysis, while its strongest results depended on radiomics and gene-expression features that may be impractical in resource-constrained settings. These studies therefore highlight the need to evaluate not only diagnostic performance but also explainability under external validation using routinely collected modalities. More fundamentally, it remains unclear whether multimodal models extract clinically transferable or invariant information from structural MRI beyond that already contained in cognitive assessments, and whether apparent fusion benefits persist under distribution shift.

\subsection{Research questions}

Addressing these gaps requires directly asking:

\begin{itemize}
    \item How does the relative contribution of imaging and clinical information change across different stages of Alzheimer's disease?
    \item To what extent do multimodal and unimodal models trained on ADNI retain their diagnostic performance when evaluated on an independent cohort?
    \item Can multimodal explainability identify imaging and clinical biomarkers that remain consistent under external validation?
\end{itemize}

\subsection{Contributions}

In a bid to tackle these questions, we develop and empirically evaluate an explainable multimodal framework that jointly represents 3D structural MRI and routinely collected clinical and demographic data for CN, MCI, and AD classification. We characterize multimodal explainability across diagnostic tasks, modalities, and independent cohorts by modeling the two modalities separately and jointly, evaluating their predictions within ADNI and under external validation on OASIS-3. We further quantify when each modality contributes, how fusion changes model behavior, and whether feature-level explanations remain consistent across cohorts, providing a systematic assessment of the relationship between diagnostic performance, modality contribution, and explanation stability under cohort variability.

The remainder of this paper is organized as follows. Section~\ref{sec:method} describes the proposed methodology and experimental design. Section~\ref{sec:results} presents diagnostic performance together with the explainability analyses on both internal and external cohorts. Section~\ref{sec:discus} discusses the implications of the findings, with particular emphasis on multimodal complementarity and cross-cohort generalization, and Section~\ref{sec:con} concludes the paper.

\section{ Materials and methods}
\label{sec:method}

\subsection{Dataset}
\label{sec:dataset}
We utilize two cohorts in this study: the Alzheimer’s Disease Neuroimaging Initiative (ADNI) for model development, training, and internal (in-distribution) evaluation, and an independent cohort, OASIS-3, reserved exclusively for external validation. We describe the ADNI cohort here and the OASIS-3 cohort.

We use data from ADNI, a longitudinal multisite study designed to develop validated biomarkers for AD detection and progression. Subjects, in this present paper, are classified into three diagnostic groups: CN, MCI, and AD. Each subject contributes a T1-weighted sMRI scan and a set of clinical and demographic variables including age, sex, education, Mini-Mental State Examination (MMSE) score, APOE4, etc. The T1-weighted sMRI scans have an original matrix size of $256 \times 256 \times 166$ voxels (X $\times$ Y $\times$ Z), with in-plane pixel spacing of $0.9 \times 0.9$ mm and a slice thickness of $1.2$ mm, corresponding to a voxel spacing of $0.9 \times 0.9 \times 1.2$ mm$^3$. Following preprocessing, all MRI volumes are standardized to $128 \times 128 \times 128$ voxels. We have provided the complete list of the structured tabular predictors in (Appendix Table B1).

\begin{figure}[!ht]
    \centering
    \includegraphics[width=1.0\linewidth]{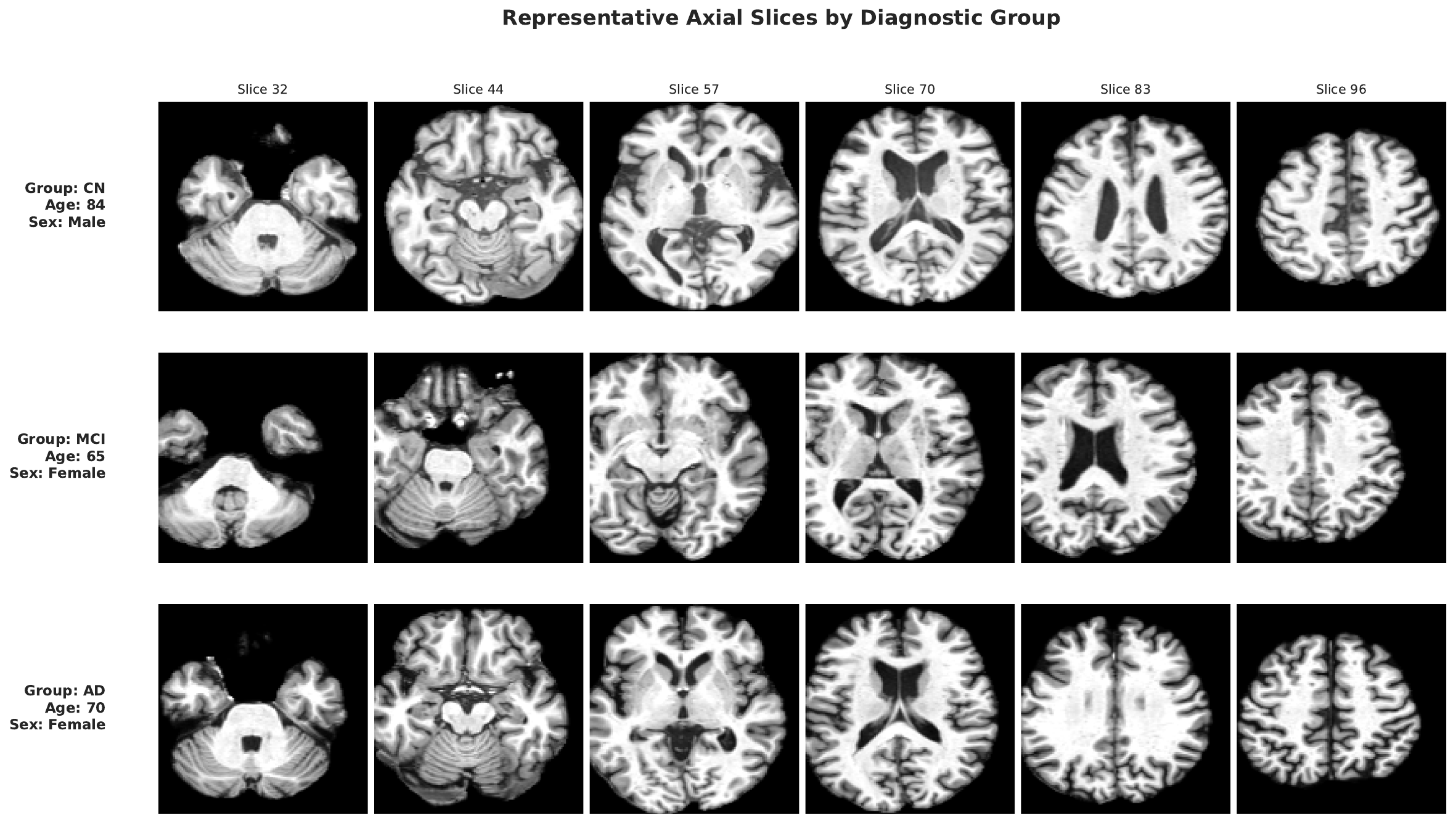}
    \caption{\textbf{Representative MRI Slices.} Representative axial slices at six anatomical levels for a cognitively normal subject (CN, age 84, male), a subject with mild cognitive impairment (MCI, age 65, female), and a subject with Alzheimer's disease (AD, age 70, female).}
    \label{fig:brain_slices_sharp}
\end{figure}

Representative axial slices across diagnostic groups are shown in Fig.~\ref{fig:brain_slices_sharp}. Dataset characteristics including age and sex distributions across diagnostic groups are provided in Appendix A.

We curate a cohort containing 6,479 3D sMRI scans with linkable electronic health records (EHRs) from the ADNI dataset, aligning each volume with its longitudinal clinical encounter using the nearest-visit matching procedure described in Appendix D. The resulting diagnostic groups are imbalanced, comprising 3,026 MCI, 2,240 CN, and 1,213 AD scans; we address this imbalance directly in our training objective in Section~\ref{sec:learning_objective}. We partition the cohort into 5174/714/591 training, validation, and test samples using an 80/10/10 percent split stratified by diagnostic group using a subject-level split strategy. The tabular feature matrix comprises $n = 7$ clinical and demographic variables per subject (Appendix Table B1), and each sMRI volume is resampled to a fixed tensor shape of $\mathbb{R}^{128 \times 128 \times 128}$ voxels (see Appendix C).

\subsection{External Validation Cohort: OASIS3}
\label{sec:oasis3}
To assess whether our framework's diagnostic performance generalizes beyond the acquisition protocol, diagnostic criteria, and population characteristics of its development cohort, we evaluate our trained models on an independent cohort drawn from the Open Access Series of Imaging Studies--3 (OASIS-3), a longitudinal, multimodal neuroimaging and clinical dataset compiled by the Washington University Knight Alzheimer Disease Research Center~\cite{lamontagne2019oasis}. OASIS-3 differs from ADNI in both recruitment and diagnostic composition: where ADNI enrolled subjects spanning the cognitively normal to established AD spectrum specifically for biomarker development, OASIS-3 originates from a predominantly preclinical, community-based cohort followed longitudinally, with diagnostic status tracked via the Clinical Dementia Rating and the Uniform Data Set neuropsychological battery. The OASIS-3 T1-weighted sMRI volumes have an image matrix size of $\mathbb{R}^{160 \times 240 \times 256}$ voxels, with an isotropic voxel spacing of $1.0 \times 1.0 \times 1.0$ mm$^3$. This difference in cohort composition and acquisition site makes OASIS-3 a meaningful test of out-of-distribution generalization.

We use the same input modalities as in model development: a 3D T1-weighted sMRI volume, preprocessed with the identical pipeline described in Appendix C, and a corresponding vector of clinical and demographic covariates per subject. Because OASIS-3 does not administer several of the ADNI-specific neuropsychological instruments (e.g., ADAS-Cog, RAVLT) used to construct our ADNI feature vector, we restrict the tabular input to a harmonized feature set comprising variables with a direct or closely corresponding equivalent in OASIS-3, applying the imputation, encoding, and scaling parameters fitted on the ADNI training set to avoid information leakage from the external cohort.

We curate 1,703 3D sMRI scans with linkable clinical assessments from OASIS-3, each mapped to one of the same three diagnostic groups used in model development, CN, MCI, and AD. The resulting cohort is markedly more imbalanced than ADNI, comprising 1,323 CN, 295 MCI, and 85 AD scans, reflecting OASIS-3's predominantly preclinical recruitment focus. We use the entire OASIS-3 cohort exclusively for external inference. The complete dataset summary across cohorts is presented in Table~\ref{tab:cohort_summary}.

\begin{table}[htbp]
\centering
\caption{Cohort and diagnostic group summary for ADNI and OASIS-3. ADNI is
used for model development (train/validation/test), while OASIS-3 is
reserved exclusively for external validation.}
\label{tab:cohort_summary}
\begin{tabular}{llrrl}
\toprule
Cohort & Diagnostic Group & N (samples) & \% of Cohort & Role \\
\midrule
\multirow{4}{*}{ADNI}
 & CN  & 2,240 & 34.6\%  & \multirow{4}{*}{Train (5,174) / Val (714) / Test (591)} \\
 & MCI & 3,026 & 46.7\%  & \\
 & AD  & 1,213 & 18.7\%  & \\
 \cmidrule(lr){2-4}
 & \textbf{Total} & \textbf{6,479} & \textbf{100\%} & \\
\midrule
\multirow{4}{*}{OASIS-3}
 & CN  & 1,323 & 77.7\% & \multirow{4}{*}{External validation only (no split)} \\
 & MCI & 295   & 17.3\% & \\
 & AD  & 85    & 5.0\%  & \\
 \cmidrule(lr){2-4}
 & \textbf{Total} & \textbf{1,703} & \textbf{100\%} & \\
\bottomrule
\end{tabular}
\end{table}

\subsection{Data Augmentation}
\label{sec:augmentation}
We applied stochastic data augmentation to the imaging modality during training; validation and test volumes receive solely the deterministic pre-processing pipeline of Appendix C. Because our pre-processing pipeline applied voxel-wise Z-score intensity normalization, we restricted augmentation to purely geometric transformations that preserve this normalized intensity distribution, avoiding intensity-domain perturbations such as contrast or noise augmentation that could re-introduce inconsistent scale artifacts. We applied random sagittal and coronal reflections with probability 0.5 each to discourage the model from memorizing a fixed hemispheric orientation, followed by a random affine transformation with probability 0.5 comprising rotation of up to $\pm 15\deg$ per axis and isotropic scaling of $\pm 10\%$ with bilinear interpolation, discouraging memorization of absolute brain position and size. All transformations are implemented using Medical Open Network for AI (MONAI)~\cite{cardoso2022monai}.

\subsection{Model Architecture}
\label{sec:model_architecture}
Structural MRI and tabular clinical variables differ in representational form: the former is a high-dimensional, spatially structured voxel grid governed by anatomical geometry, while the latter is a low-dimensional, heterogeneous vector of scalar and categorical clinical measurements with no inherent spatial structure. Learning a single shared representation directly from the concatenated raw inputs, an early fusion strategy, would require one encoder to simultaneously model two disparate generative structures, and risks the numerically larger, higher-variance imaging modality dominating early gradient signal and representational capacity~\cite{baltruvsaitis2018multimodal}. We instead adopt a late, representation-level fusion strategy: each modality is first mapped, through an encoder whose inductive priors match its input structure, a 3D convolutional neural network (3D CNN) for the volumetric MRI and a multilayer perceptron (MLP) for the tabular vector, into a compact, modality-specific embedding, and only these already-abstracted representations are combined~\cite{bengio2013representation}. This design keeps the representation learning problem for each modality tractable and optimizable, and, because fusion operates on per-modality summaries rather than raw signal, preserves a direct correspondence between each embedding and its originating modality. That correspondence is what affords modality-specific attribution methods we use in Section~\ref{sec:results}, Grad-CAM++ and Score-CAM on the imaging branch and SHAP on the tabular branch: an early-fusion architecture would entangle modality contributions from the first layer onward, thus, undermining the clean separation explainability analysis depends on. Late fusion additionally decouples the choice of combination mechanism from the choice of encoder, letting us hold both modality-specific branches fixed while varying only the fusion mechanism.

\begin{figure}[!htb]
    \centering
    \includegraphics[
    width=\linewidth]{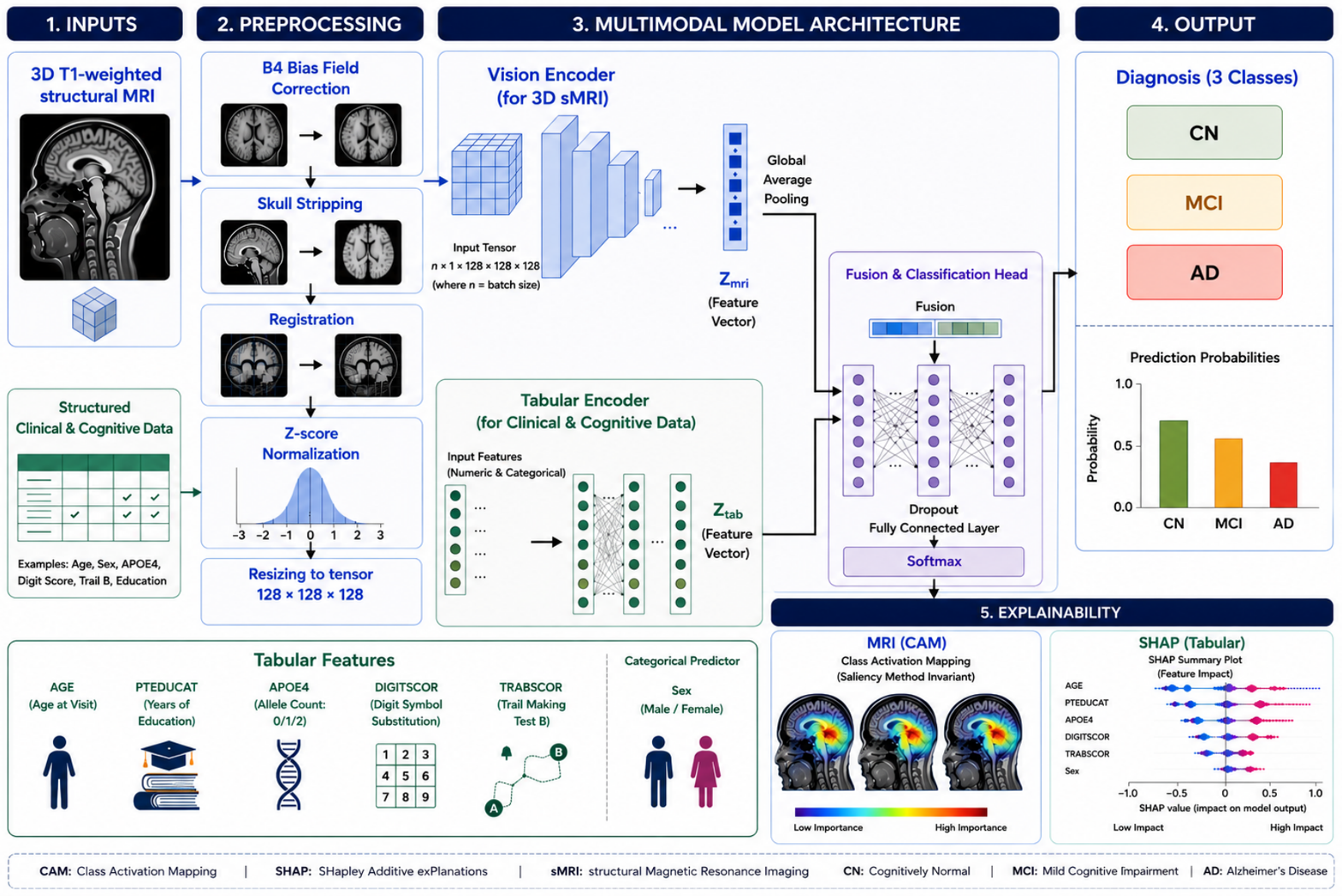}
    \caption{\textbf{Multimodal Modelling Framework.}  Overview of the multimodal modelling framework. (1)~Data acquisition: structural T1-weighted MRI and clinical and demographic tabular data. (2)~Preprocessing: standardized pipeline including orientation correction, bias field correction, skull stripping, registration,intensity normalization, and spatial resampling and cropping. (3)~Model architecture: parallel 3D Vision Encoder and Tabular Encoder branches with feature-level fusion and a fully connected classifier. (4)~Explainability: Class activation maps for the imaging encoder and SHAP feature attributions for the tabular encoder.}
    \label{fig:Modelling_Framework}
\end{figure}

We investigate two such fusion strategies, illustrated in Fig.~\ref{fig:Modelling_Framework}, against unimodal training setups.

\subsubsection{Vision Encoder} 
We encode each pre-processed 3D sMRI volume, represented as a single-channel input tensor of shape $\mathbb{R}^{1 \times 128 \times 128 \times 128}$, using a 3D DenseNet-121 backbone~\cite{huang2017densely}, adapted to single-channel volumetric input. We repurpose the network's final fully connected layer as a linear projection into a $d_{\text{img}} = 128$-dimensional embedding space so that the imaging branch performs representation extraction only, deferring the classification decision to the fusion head. We apply dropout with $p = 0.2$ within the backbone to regularize the network's parameter count relative to the size of the training cohort.

\subsubsection{Tabular Encoder}
We encode the $n = 7$ clinical and demographic features with a lightweight multilayer perceptron: a linear layer projects the input to a 128-dimensional hidden representation, followed by batch normalization, a ReLU non-linearity, and dropout with probability $p = 0.3$, before a second linear layer projects to the final tabular embedding $\mathbf{f}_{\text{tab}} \in \mathbb{R}^{d_{\text{tab}}}$, where $d_{\text{tab}} = 64$ denotes the embedding dimensionality.

\subsubsection{Fusion strategies}
We evaluate two mechanisms for combining the imaging embedding $\mathbf{f}_{\text{img}} \in \mathbb{R}^{d_{\text{img}}}$ and the tabular embedding $\mathbf{f}_{\text{tab}} \in \mathbb{R}^{d_{\text{tab}}}$ produced by the two encoders above. Concatenation fusion directly concatenates the two embeddings, writing $[\mathbf{f}_{\text{img}} \,\|\, \mathbf{f}_{\text{tab}}]$ where $\|$ denotes vector concatenation, and passes the result to a shared classification head, as formalized in equation~\ref{eq:concat_fusion}.

Cross-attention fusion instead projects both embeddings into a shared $d_{\text{model}}$-dimensional space, with $d_{\text{model}} = 64$ denoting the common projection dimensionality, and treats the imaging embedding as a query attending over the tabular embedding as key and value:
\begin{equation}
  \mathbf{q} = W_{\text{img}}\mathbf{f}_{\text{img}}, \qquad \mathbf{k} = \mathbf{v} = W_{\text{tab}}\mathbf{f}_{\text{tab}},
  \label{eq:qkv_proj}
\end{equation}
where $\mathbf{q} \in \mathbb{R}^{d_{\text{model}}}$ is the query vector derived from the imaging embedding; $\mathbf{k} = \mathbf{v} \in \mathbb{R}^{d_{\text{model}}}$ are the key and value vectors, identical here since both derive from the single tabular embedding; and $W_{\text{img}} \in \mathbb{R}^{d_{\text{model}} \times d_{\text{img}}}$ and $W_{\text{tab}} \in \mathbb{R}^{d_{\text{model}} \times d_{\text{tab}}}$ are learned projection matrices mapping each modality's embedding into the shared space. We then compute
\begin{equation}
  \mathbf{z} = \text{LayerNorm}\!\left(\mathbf{q} + \text{MultiHeadAttn}(\mathbf{q}, \mathbf{k}, \mathbf{v})\right),
  \label{eq:cross_attn}
\end{equation}
where $\text{MultiHeadAttn}(\cdot)$ denotes standard multi-head scaled dot-product attention~\cite{vaswani2017attention}, $\text{LayerNorm}(\cdot)$ denotes layer normalization, and $\mathbf{z} \in \mathbb{R}^{d_{\text{model}}}$ is the resulting fused representation, produced via a residual connection from the query prior to normalization. We use four attention heads, with dropout probability $p = 0.1$ applied within the attention computation. This formulation allows the model to learn a sample-specific, soft re-weighting of tabular information conditioned on the imaging representation, rather than combining the two modalities at a fixed ratio as concatenation fusion does.

\textbf{Shared classification head.} Regardless of fusion strategy, the fused representation, $\mathbb{R}^{d_{\text{img}}+d_{\text{tab}}} = \mathbb{R}^{192}$ under concatenation or $\mathbb{R}^{d_{\text{model}}} = \mathbb{R}^{64}$ under cross-attention, is passed through an identical fully connected head consisting of a linear layer to 64 units, batch normalization, ReLU, dropout with probability $p = 0.4$, and a final linear layer to the diagnostic class logits, giving the general form:
\begin{equation}
  \hat{y} = \text{softmax}\!\left( W \cdot [\mathbf{f}_{\text{img}} \,\|\, \mathbf{f}_{\text{tab}}] + b \right),
  \label{eq:concat_fusion}
\end{equation}
where $\hat{y} \in [0,1]^{C}$ is the predicted probability vector over the $C$ diagnostic classes, $\text{softmax}(\cdot)$ is the standard softmax normalization, and $W$ and $b$ are the weight matrix and bias vector of the head's final linear layer. Under cross-attention fusion, the bracketed term $[\mathbf{f}_{\text{img}} \,\|\, \mathbf{f}_{\text{tab}}]$ is replaced by $\mathbf{z}$, with $W$ and $b$ resized accordingly.

\subsubsection{Unimodal Learning} To isolate the contribution of each modality, we train two unimodal classifiers alongside the fused models. The tabular-only classifier applies a dropout layer and a linear classification head directly to the tabular branch embedding. The vision-only classifier uses the identical 3D DenseNet-121 backbone as the imaging branch of the multimodal model, with a classification head appended in place of the embedding projection. Using matched encoders across the unimodal and multimodal settings ensures that any performance difference we observe is attributable to fusion rather than to differences in per-modality representational capacity.

\subsection{Learning Objective}
\label{sec:learning_objective}
We train each model under one of four classification setups, indexed $t \in \{0, 1, 2, 3\}$. The default setup, $t = 0$, is the full three-class discrimination task distinguishing CN, MCI, and AD. The remaining three setups isolate individual pairwise transitions along the diagnostic spectrum: $t = 1$ contrasts CN against AD, $t = 2$ contrasts CN against MCI, and $t = 3$ contrasts MCI against AD.

To counteract the class imbalance in our cohort, we weight each class's contribution to the loss using the effective number of samples framework~\cite{cui2019class}. For a class $c$ with $n_c$ training examples, we compute an effective sample count
\begin{equation}
    E_{n_c} = \frac{1 - \beta^{n_c}}{1 - \beta}, \qquad \beta \in [0,1),
    \label{eq:effective_num}
\end{equation}
and derive the class weight as the inverse effective number, normalized across the $C$ classes present in the current setup:
\begin{equation}
    w_c = \frac{1/E_{n_c}}{\sum_{j=1}^{C} 1/E_{n_j}} \cdot C.
    \label{eq:class_weight}
\end{equation}
We set $\beta = 0.99$ and recompute $\{w_c\}$ separately for each of the four setups, using only the class counts of the diagnostic groups retained under that setup, so that the binary setups are reweighted independently of the full three-class distribution.

For $t = 0$, we optimize the class-weighted categorical cross-entropy loss over the three-way softmax output of Equation~\ref{eq:concat_fusion}:
\begin{equation}
    \mathcal{L}_{\text{CE}} = -\sum_{c=1}^{3} w_c\, y_c \log(\hat{y}_c),
    \label{eq:ce_loss}
\end{equation}
where $y_c \in \{0,1\}$ is the one-hot ground-truth indicator and $\hat{y}_c$ the predicted class probability. For $t \in \{1,2,3\}$, we restrict the training and evaluation data to the two relevant diagnostic groups and optimize the corresponding class-weighted binary cross-entropy loss:
\begin{equation}
    \mathcal{L}_{\text{BCE}} = -\left[w_1\, y \log(\hat{y}) + w_0\, (1-y)\log(1-\hat{y})\right],
    \label{eq:bce_loss}
\end{equation}
with $y \in \{0,1\}$ the binary ground-truth label and $\hat{y}$ the predicted positive-class probability. This four-setup formulation lets us contrast the model's discriminability across the full diagnostic spectrum against its discriminability at each individual transition along the disease trajectory, isolating the stages where the fused representation is most and least reliable and directly addressing the differential-diagnosis challenge raised in Section~\ref{sec:intro}.

\subsection{Performance Evaluation}
We assess diagnostic performance on the held-out test set using three complementary metrics: overall classification accuracy, micro-averaged F1 score, and one-vs-rest macro-averaged AUC--ROC. For each metric, we report 95\% confidence intervals estimated via stratified bootstrap resampling of the test set with 1,000 replicates, taking the 2.5th and 97.5th percentiles of the resulting distribution as the interval bounds. We apply this protocol uniformly across all four classification setups and across all unimodal and multimodal model variants, enabling direct, uncertainty-aware comparison. Algorithmic transparency is evaluated separately, at both global and local levels, using dual attribution systems whose cross-method agreement we report in Section~\ref{sec:results}.

\subsection{Experimental Setup}
We train all models on a single NVIDIA L4 Graphics Processing Unit (GPU) with 48\,GB of Video Random Access Memory (VRAM) using the Adam optimizer with a learning rate of $1\times10^{-4}$ and a weight decay of $1\times10^{-4}$, selected heuristically and held fixed across all model variants. We use a batch size of 32 for the tabular-only classifier and a batch size of 16 for the vision-only and multimodal models, reflecting the memory footprint of full 3D volumetric input relative to tabular features. All models are trained for a maximum of 50 epochs, with early stopping based on validation loss and a patience of 10 epochs. We compute Grad-CAM++ and Score-CAM attributions from the final convolutional layer of the imaging branch using the \texttt{grad-cam} package (version 1.5.5)~\cite{jacobgilpytorchcam}, and SHAP values for the tabular branch using the \texttt{shap} package (version 0.52.0). All experiments are implemented in Python 3.12.13 using PyTorch 2.13.0+cu130, with TorchVision 0.28.0+cu130 and TorchIO 1.2.1 for model and volumetric image processing, respectively.

\section{Results}
\label{sec:results}
We evaluated diagnostic performance first to establish the prediction regimes in which the explanations were obtained, and then examined how model attributions behaved across modalities and cohorts. We used the same modality-specific explanation framework throughout: SHAP and Integrated Gradients for tabular inputs, and Grad-CAM++ with Score-CAM for structural MRI. This allowed us to distinguish consistency between explanation methods from changes in model explanations associated with fusion and external cohort variability.

 \subsection{Diagnostic performance}

 \begin{table*}[!ht]
\centering
\caption{
Test-set performance across ADNI classification tasks.
Values represent AUC--ROC and Micro F1 scores with 95\% bootstrap confidence intervals.
\textbf{Bold} = best; \textcolor{blue}{blue} = second-best (per task, per column).
}
\label{tab:adni_multimodal_results}
\small
\setlength{\tabcolsep}{5pt}
\resizebox{\textwidth}{!}{%
\begin{tabular}{llcc}
\toprule
\textbf{Task} &
\textbf{Model} &
\textbf{AUC--ROC (95\% CI)} &
\textbf{Micro F1 (95\% CI)} \\
\midrule
\multirow{4}{*}{CN vs MCI vs AD}
& Tabular-only
& \best{0.879} (0.858--0.899)
& \best{0.760} (0.721--0.795) \\
& Vision-only
& 0.764 (0.736--0.791)
& 0.567 (0.524--0.607) \\
& Multimodal (Concat)
& 0.850 (0.826--0.873)
& 0.653 (0.614--0.692) \\
& Multimodal (Cross-Attention)
& \second{0.855} (0.834--0.878)
& \second{0.677} (0.640--0.714) \\
\midrule
\multirow{4}{*}{CN vs AD}
& Tabular-only
& \second{0.999} (0.996--1.000)
& \best{0.975} (0.957--0.991) \\
& Vision-only
& 0.939 (0.912--0.961)
& 0.837 (0.800--0.877) \\
& Multimodal (Concat)
& \best{1.000} (0.999--1.000)
& \second{0.963} (0.939--0.982) \\
& Multimodal (Cross-Attention)
& \second{0.999} (0.997--1.000)
& \best{0.975} (0.957--0.991) \\
\midrule
\multirow{4}{*}{CN vs MCI}
& Tabular-only
& \best{0.903} (0.877--0.927)
& \second{0.800} (0.768--0.835) \\
& Vision-only
& 0.754 (0.711--0.792)
& 0.649 (0.607--0.687) \\
& Multimodal (Concat)
& \second{0.886} (0.859--0.912)
& \best{0.814} (0.779--0.848) \\
& Multimodal (Cross-Attention)
& 0.864 (0.832--0.893)
& 0.766 (0.731--0.800) \\
\midrule
\multirow{4}{*}{MCI vs AD}
& Tabular-only
& \second{0.837} (0.787--0.882)
& \second{0.697} (0.640--0.754) \\
& Vision-only
& 0.601 (0.532--0.661)
& 0.629 (0.564--0.689) \\
& Multimodal (Concat)
& 0.835 (0.784--0.879)
& 0.689 (0.633--0.742) \\
& Multimodal (Cross-Attention)
& \best{0.861} (0.816--0.902)
& \best{0.837} (0.788--0.879) \\
\bottomrule
\end{tabular}%
}
\end{table*}

\begin{table*}[!ht]
\centering
\caption{
Test-set performance across OASIS3 classification tasks.
Values represent AUC--ROC and Micro F1 scores with 95\% bootstrap confidence intervals.
\textbf{Bold} = best; \textcolor{blue}{blue} = second-best (per task, per column).
}
\label{tab:oasis3_multimodal_results}
\small
\setlength{\tabcolsep}{5pt}
\resizebox{\textwidth}{!}{%
\begin{tabular}{llcc}
\toprule
\textbf{Task} &
\textbf{Model} &
\textbf{AUC--ROC (95\% CI)} &
\textbf{Micro F1 (95\% CI)} \\
\midrule
\multirow{4}{*}{CN vs MCI vs AD}
& Tabular-only
& 0.798 (0.780--0.815)
& 0.622 (0.597--0.646) \\
& Vision-only
& \best{0.910} (0.899--0.921)
& \second{0.805} (0.787--0.823) \\
& Multimodal (Concat)
& \second{0.909} (0.896--0.923)
& \best{0.813} (0.793--0.834) \\
& Multimodal (Cross-Attention)
& 0.874 (0.859--0.889)
& 0.746 (0.723--0.770) \\
\midrule
\multirow{4}{*}{CN vs AD}
& Tabular-only
& 0.875 (0.846--0.901)
& \second{0.906} (0.890--0.921) \\
& Vision-only
& \second{0.876} (0.829--0.915)
& \best{0.954} (0.942--0.965) \\
& Multimodal (Concat)
& 0.856 (0.824--0.887)
& 0.898 (0.882--0.914) \\
& Multimodal (Cross-Attention)
& \best{0.886} (0.858--0.911)
& 0.904 (0.889--0.920) \\
\midrule
\multirow{4}{*}{CN vs MCI}
& Tabular-only
& 0.563 (0.480--0.640)
& 0.667 (0.642--0.692) \\
& Vision-only
& \best{0.671} (0.631--0.708)
& \best{0.815} (0.795--0.833) \\
& Multimodal (Concat)
& 0.565 (0.471--0.654)
& \second{0.750} (0.727--0.773) \\
& Multimodal (Cross-Attention)
& \second{0.574} (0.490--0.654)
& 0.728 (0.704--0.752) \\
\midrule
\multirow{4}{*}{MCI vs AD}
& Tabular-only
& \second{0.791} (0.728--0.846)
& 0.407 (0.348--0.467) \\
& Vision-only
& 0.567 (0.501--0.635)
& \best{0.776} (0.734--0.818) \\
& Multimodal (Concat)
& 0.788 (0.726--0.847)
& \second{0.504} (0.444--0.567) \\
& Multimodal (Cross-Attention)
& \best{0.806} (0.746--0.860)
& 0.411 (0.352--0.470) \\
\bottomrule
\end{tabular}%
}
\end{table*}

On ADNI, the tabular-only model achieved the highest three-class AUC--ROC (0.879, 95\% CI 0.858--0.899), followed by cross-attention fusion (0.855, 0.834--0.878), concatenation fusion (0.850, 0.826--0.873), and the vision-only model (0.764, 0.736--0.791) (Table~\ref{tab:adni_multimodal_results}). The pairwise tasks revealed a marked dependence on diagnostic transition. CN versus AD was highly distinguishable across models (AUC--ROC 0.939--1.000), and CN versus MCI was likewise well discriminated by the clinical and fused models, led by the tabular-only (0.903, 0.877--0.927) and concatenation-fusion (0.886, 0.859--0.912) models, with the vision-only model trailing well behind at 0.754 (0.711--0.792). For MCI versus AD, cross-attention achieved the highest AUC--ROC (0.861, 0.816--0.902), while the vision-only model achieved only 0.601 (0.532--0.661). These results establish substantial variation in the information available to each modality across diagnostic stages, with clinical variables carrying the CN-versus-MCI transition and cross-attention fusion carrying MCI-versus-AD.

External evaluation on OASIS-3 produced a varied performance profile. The vision-only model achieved the highest three-class AUC--ROC (0.910, 0.899--0.921), closely followed by concatenation fusion (0.909, 0.896--0.923), whereas the tabular-only and cross-attention models achieved 0.798 (0.780--0.815) and 0.874 (0.859--0.889), respectively (Table~\ref{tab:oasis3_multimodal_results}). CN versus MCI remained difficult, with AUC--ROC ranging from 0.563 to 0.671, while cross-attention achieved the highest MCI versus AD AUC--ROC (0.806, 0.746--0.860). Thus, neither modality nor fusion strategy retained a uniform performance ranking across cohorts, providing a relevant setting in which to examine whether the associated explanations also changed.

\subsection{Tabular explainability across cohorts}
\begin{figure}[!ht]
    \centering
    \includegraphics[width=0.450\linewidth]{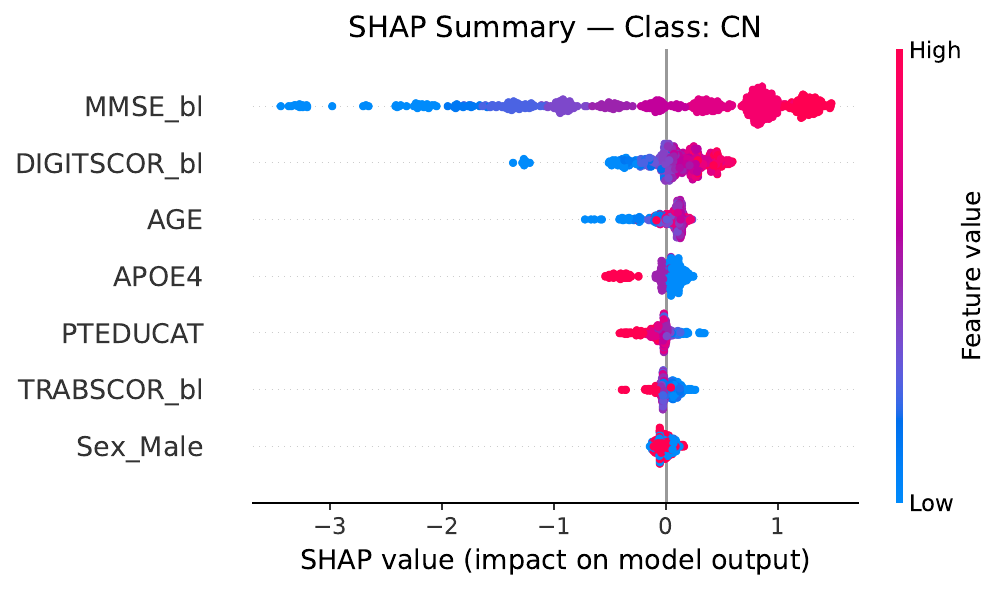}
    \includegraphics[width=0.450\linewidth]{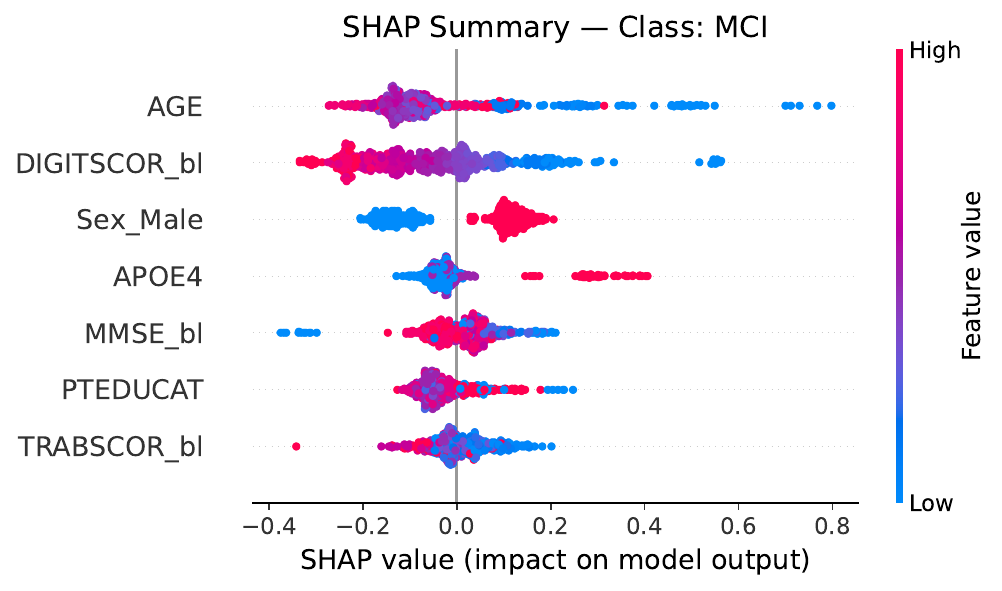}
    \includegraphics[width=0.450\linewidth]{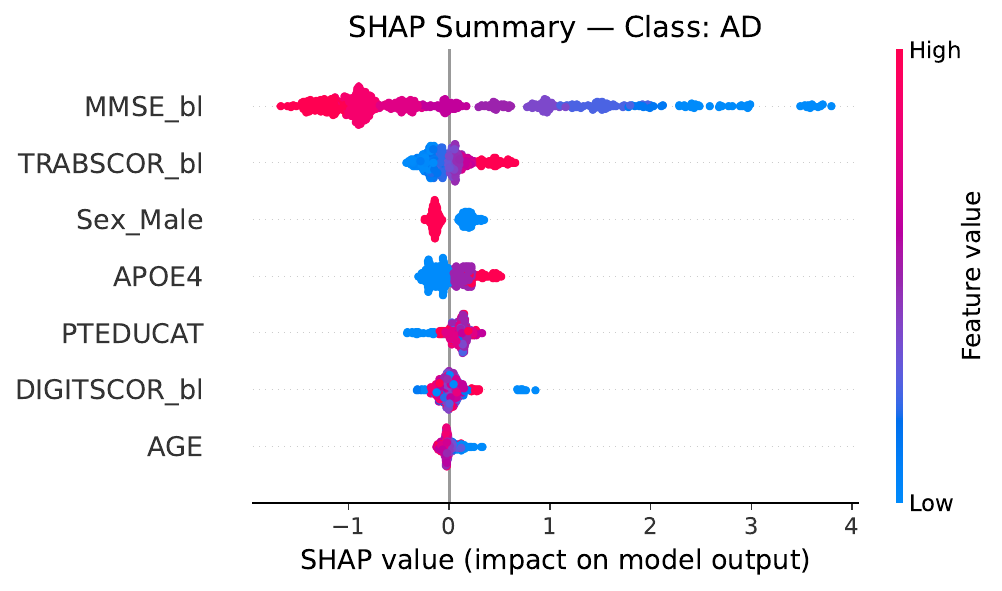}
    \caption{\textbf{SHAP Summary by Diagnostic Class.} SHAP summary plots showing feature-level attributions for the CN, MCI, and AD target classes.}
    \label{fig:shap_summary_AD}
\end{figure}

On ADNI, SHAP identified baseline MMSE as the most prominent feature across the diagnostic classes, with additional contributions from processing speed, executive function, age, sex, APOE4 status, and education (Fig.~\ref{fig:shap_summary_AD}). 
At the class level, high baseline MMSE scores exerted a positive impact on the model output for the CN class, whereas lower scores showed a strong negative impact. 
Conversely, for the AD class, low baseline MMSE scores demonstrated a pronounced positive impact on the model prediction, extending up to a SHAP value of approximately 7. 
For the MCI class, the impact of baseline MMSE was tightly clustered around zero, where instead higher biological age and lower baseline digit test scores displayed broader distributions of positive impact.
The corresponding global importance profiles from SHAP and Integrated Gradients showed strong agreement, with a Spearman rank correlation of $\rho=0.94$. This agreement indicates that the dominant feature ordering was not specific to a single attribution method.

The same analysis on OASIS-3 produced an identical global ranking between SHAP and Integrated Gradients ($\rho=0.96$; Fig.~\ref{fig:shap_importance_heatmap_by_class_oasis3}). Baseline MMSE remained the dominant feature in the global importance profile, providing a fairly consistent signal across cohorts. At the class level, however, the relative ordering of the remaining features varied between ADNI and OASIS-3. Thus, the explanations showed a stable dominant cognitive signal but were not invariant at the level of individual feature rankings across diagnostic classes.

\begin{figure}[!ht]
    \centering
    \includegraphics[width=0.450\linewidth]{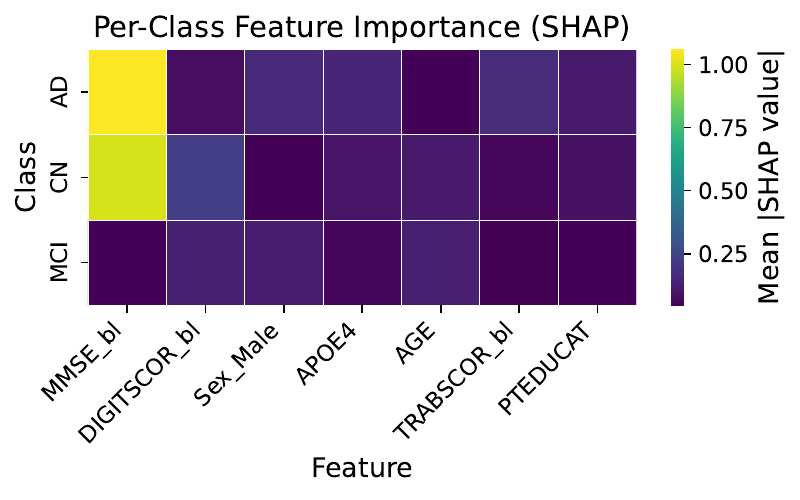}
    \includegraphics[width=0.450\linewidth]{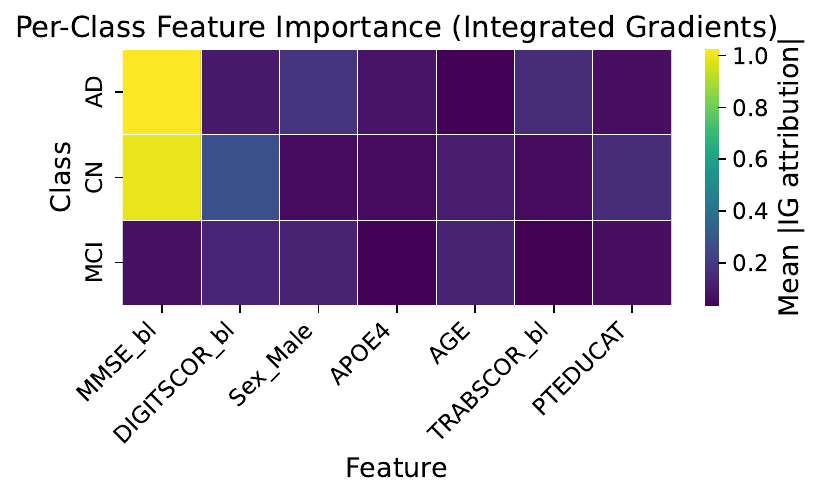}
    \caption{\textbf{Global Feature Importance.} Per-class global feature importance from SHAP (left) and Integrated Gradients (right).}

    \label{fig:shap_importance_heatmap_by_class}
\end{figure}

\subsection{Visual explainability across cohorts}
\begin{figure}[!ht]
    \centering
    \includegraphics[width=1.0\linewidth]{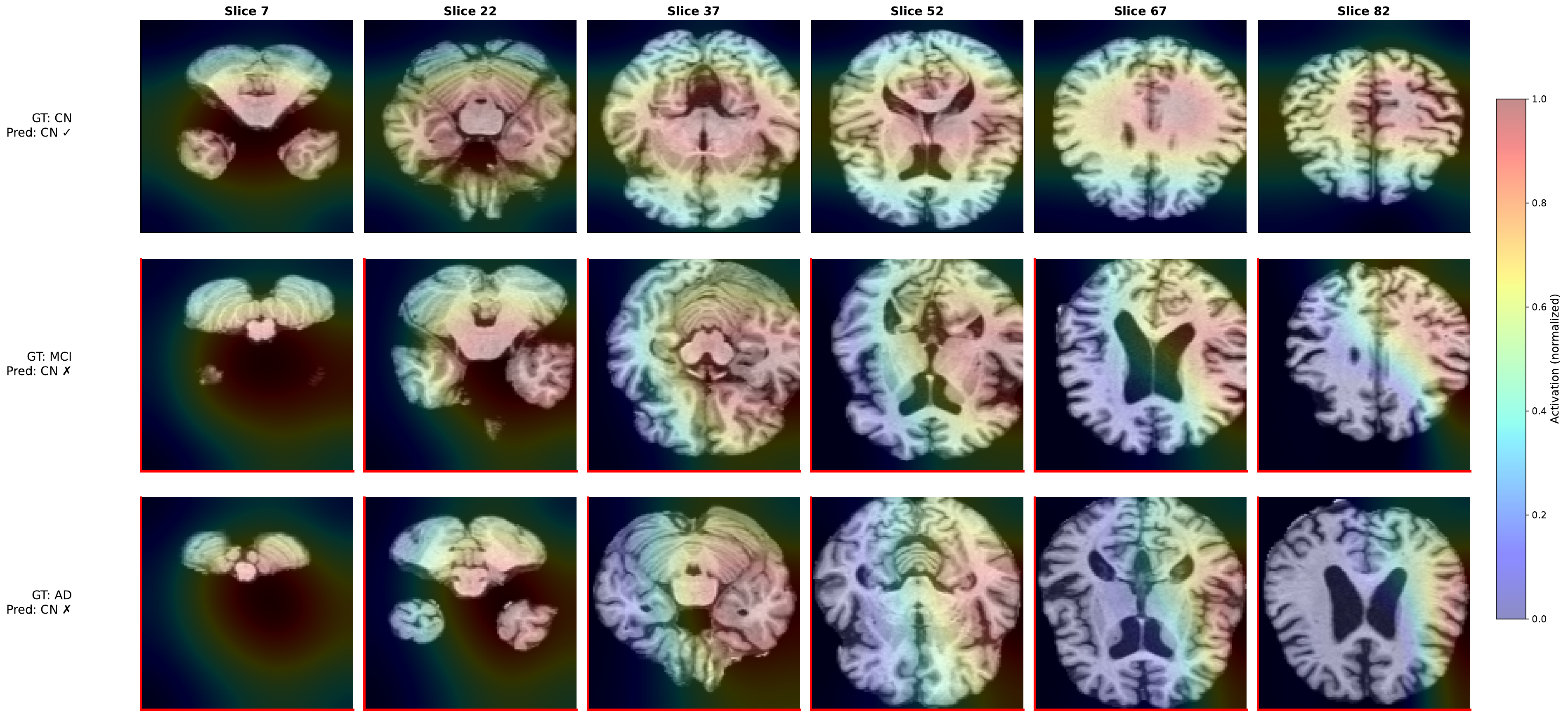}
    \caption{\textbf{Vision-Only Grad-CAM++ Attributions.} Grad-CAM++ attribution maps for the vision-only model across six representative axial slices for CN, MCI, and AD testset subjects. The CN subject is correctly classified, whereas the MCI and AD subjects are misclassified as CN (red outlines). Warmer colors indicate higher normalized activation.}
    \label{fig:gradcampp_grid}
\end{figure}

\begin{figure}[!htp]
    \centering
    \includegraphics[width=1.0\linewidth]{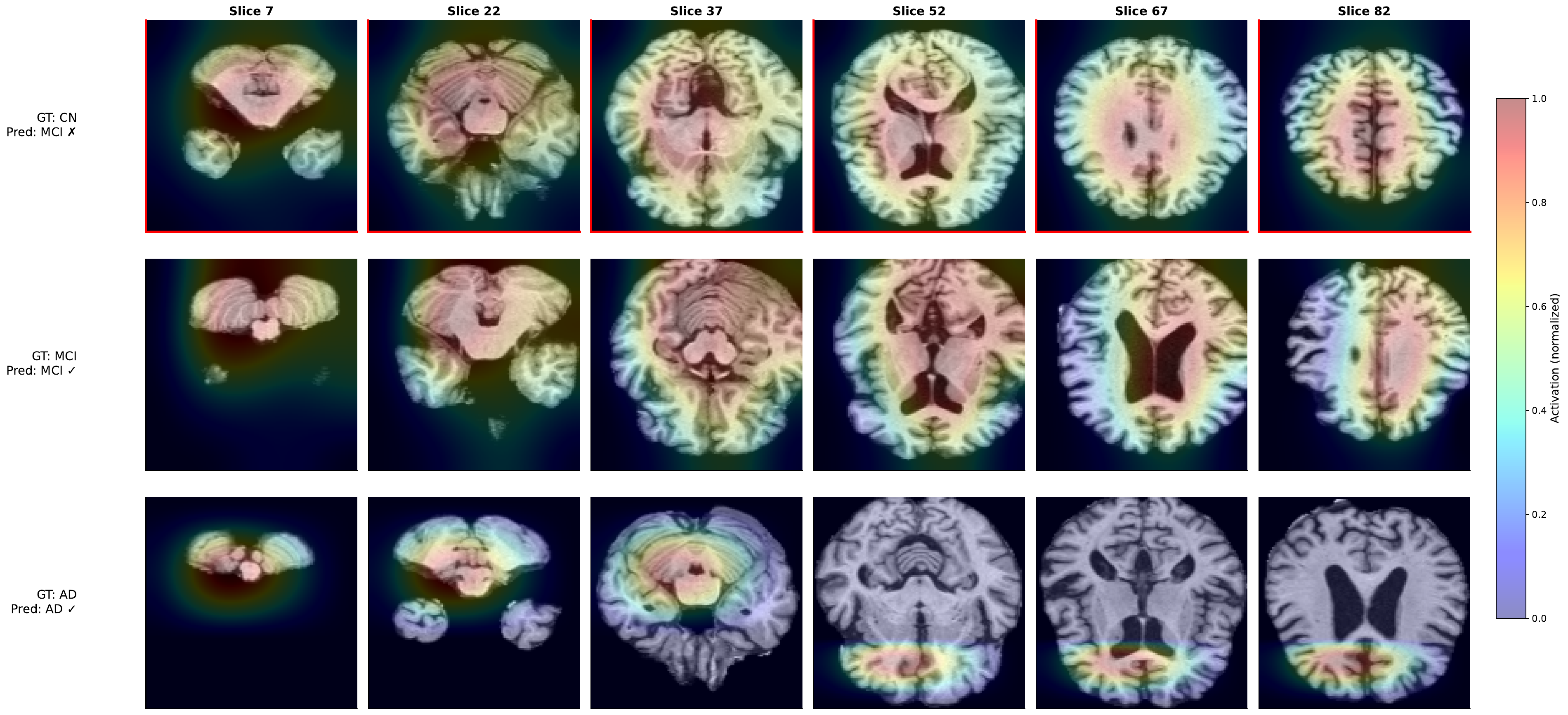}
    \caption{\textbf{Concatenation-fusion Grad-CAM++ Attributions.} Grad-CAM++ attribution maps for the Concatenation-fusion multimodal model on the same ADNI testset subjects shown in Fig.~\ref{fig:gradcampp_grid}. The CN subject is misclassified as MCI, while the MCI and AD subjects are correctly classified.}
    \label{fig:gradcampp_grid_multimodal_concat}
\end{figure}

\begin{figure}[!ht]
    \centering
    \includegraphics[width=1.0\linewidth]{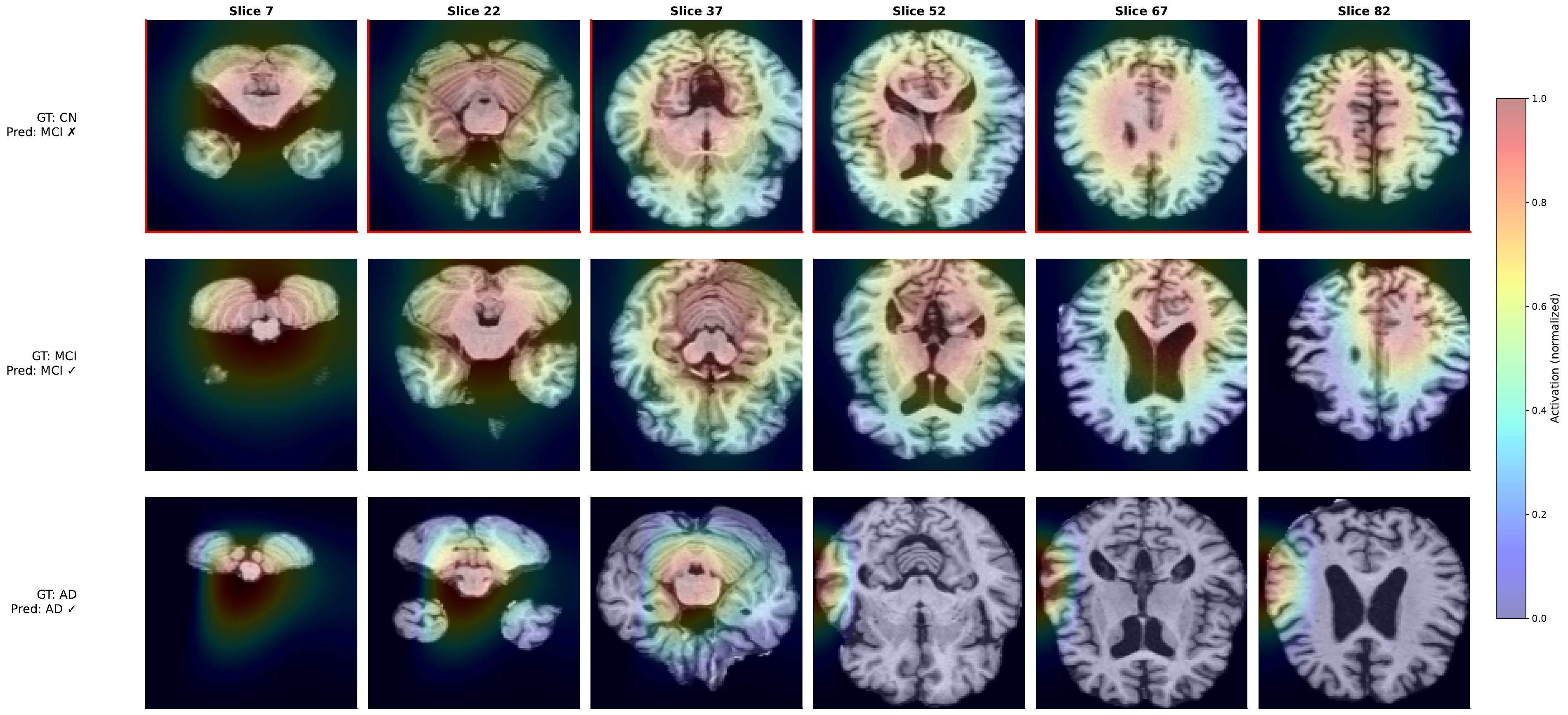}
    \caption{\textbf{Cross-Attention Grad-CAM++ Attributions.} Grad-CAM++ attribution maps for the cross-attention-fusion multimodal model on the same ADNI testset subjects shown in Figs~\ref{fig:gradcampp_grid} and~\ref{fig:gradcampp_grid_multimodal_concat}. The CN subject is misclassified as MCI, while the MCI and AD subjects are correctly classified.}
    \label{fig:gradcampp_grid_multimodal_cross_attn}
\end{figure}

On ADNI, Grad-CAM++ showed that the vision-only model correctly classified the representative CN subject but assigned both the MCI and AD subjects to CN (Fig.~\ref{fig:gradcampp_grid}). For the same subjects, concatenation and cross-attention fusion correctly classified the MCI and AD subjects but assigned the CN subject to MCI (Figs.~\ref{fig:gradcampp_grid_multimodal_concat} and~\ref{fig:gradcampp_grid_multimodal_cross_attn}). We observed that Score-CAM produced more coarse saliency maps compared to Grad-CAM++ for these subjects (Appendix Fig. E3).

The corresponding OASIS-3 examples showed a different pattern. The vision-only model correctly classified the CN subject but assigned the MCI subject to CN and the AD subject to MCI, whereas both multimodal models assigned the CN and MCI subjects to AD and correctly classified the AD subject (Appendix Figs F8-F11). The visual explanations therefore changed together with the model predictions across cohorts. These examples demonstrate that image-level explanations can be sensitive to the cohort in which the model is applied, although the representative nature of these cases does not permit a cohort-level quantitative assessment of spatial explanation stability.

\subsection{Multimodal Explainability}

\begin{figure}[!ht]
    \centering
    \includegraphics[width=1.0\linewidth,height=0.7\textheight, keepaspectratio]{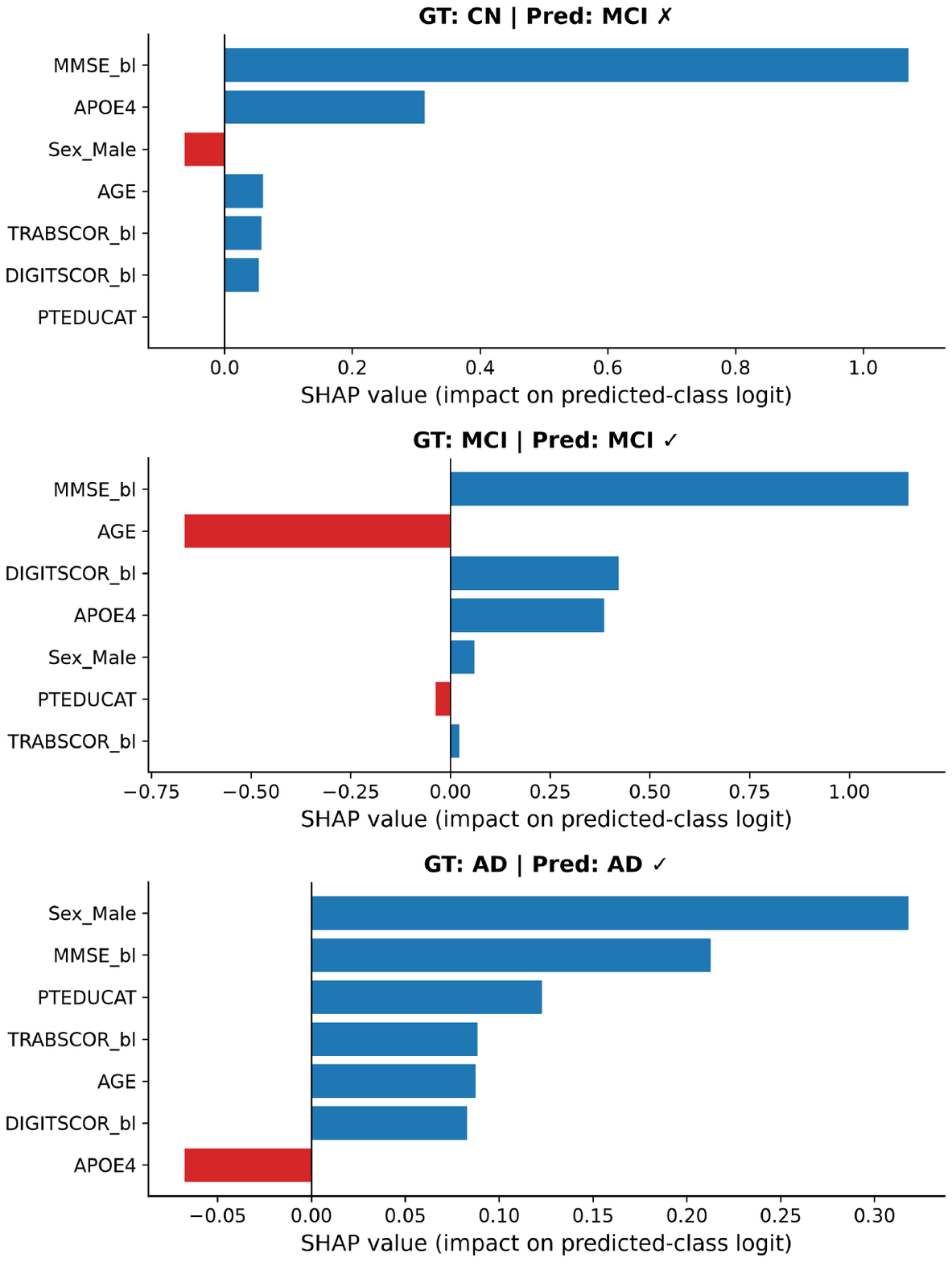}
    \caption{\textbf{Concatenation-fusion SHAP Attributions.} SHAP feature attributions for the predicted-class from the tabular branch of the Concatenation-fusion model for the CN, MCI, and AD testset subjects shown in Fig.~\ref{fig:gradcampp_grid_multimodal_concat}. Blue bars indicate positive and red bars negative contributions to the predicted-class.}
    \label{fig:tabular_shap_per_class_concat}
\end{figure}

\begin{figure}[!ht]
    \centering
    \includegraphics[width=1.0\linewidth,height=0.7\textheight, keepaspectratio]{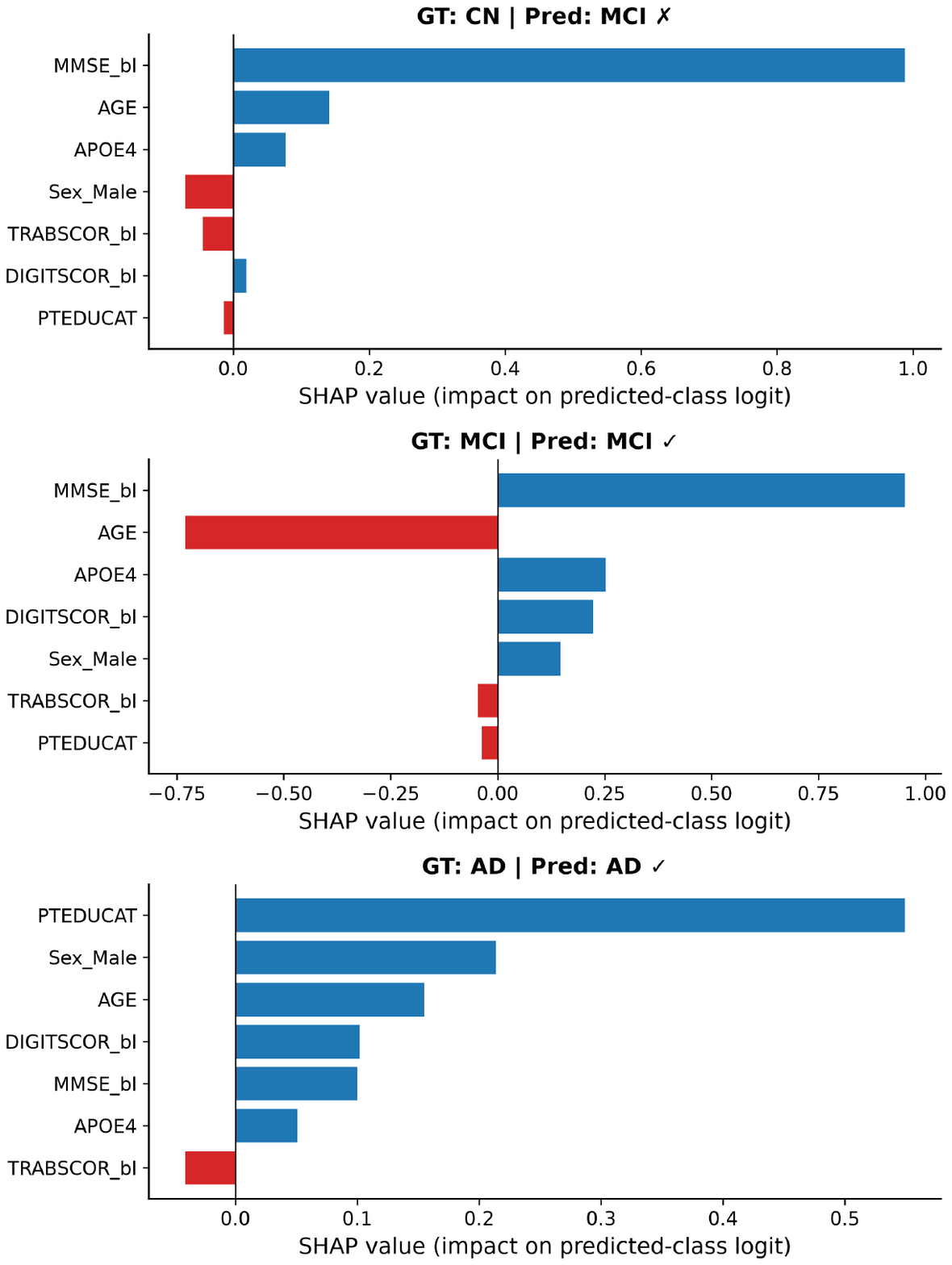}
    \caption{\textbf{Cross-Attention SHAP Attributions.} SHAP feature attributions for the predicted-class using the tabular branch of the cross-attention-fusion model for the CN, MCI, and AD subjects shown in Fig.~\ref{fig:gradcampp_grid_multimodal_cross_attn}. Blue bars indicate positive and red bars negative contributions to the predicted-class.}
    \label{fig:tabular_shap_per_class_cross_attn}
\end{figure}

The multimodal explanations further showed that fusion altered the clinical evidence associated with individual predictions. For the ADNI examples, SHAP attributions for the tabular branch differed between concatenation and cross-attention despite using the same underlying clinical inputs (Figs~\ref{fig:tabular_shap_per_class_concat} and~\ref{fig:tabular_shap_per_class_cross_attn}). This indicates that the fusion mechanism affected not only the final prediction but also the relative contribution assigned to the clinical variables.

The same behavior was observed under external validation. In OASIS-3, the multimodal models produced different predictions from the vision-only model for the same subjects, while their tabular SHAP profiles also changed with the predicted class (Appendix Figs. F10 and F14). Taken together, the results show that explanation behavior is coupled to both modality and fusion strategy: the dominant global cognitive signal was reproducible across cohorts, whereas local feature contributions and image-level attribution patterns changed with the model and cohort.

\section{Discussion}
\label{sec:discus}
\subsection{Model performance and generalization}

Our primary finding is not that one architecture consistently outperformed the others, but that the relative value of each modality changed across diagnostic tasks and cohorts (Tables~\ref{tab:adni_multimodal_results} and~\ref{tab:oasis3_multimodal_results}). Clinical variables carried the CN-versus-MCI transition and the ADNI three-class task, whereas cross-attention fusion carried MCI-versus-AD in both cohorts, and the vision-only model became the strongest three-class predictor only after the shift to OASIS-3. This reversal, imaging moving from the weakest to the strongest unimodal predictor between cohorts, is the clearest evidence that no single modality or fusion strategy is intrinsically superior; each performs well only within a specific diagnostic-task and cohort context. These results argue against interpreting multimodal fusion as uniformly additive and instead establish the prediction context in which explanation behavior must be evaluated.

This performance shift is important precisely because it reframes how the explanations in Section~\ref{sec:results} should be read. A model can retain apparently plausible attribution patterns while operating under a different predictive regime, and conversely, a change in attribution does not by itself establish model failure. Because the same architecture and modality combination behaved differently across cohorts, within-cohort explanation alone is insufficient for assessing the robustness of model reasoning. External validation should therefore extend beyond discrimination metrics to include examination of whether the evidence used by the model remains comparable.

\subsection{Explainability across modalities and cohorts}

The tabular analyses provide the clearest evidence for partial explanation consistency across cohorts. MMSE remained the dominant global feature in both ADNI and OASIS-3, and SHAP and Integrated Gradients showed strong within-cohort agreement ($\rho=0.94$ and $\rho=0.96$, respectively). At the same time, the class-specific ordering of secondary features changed across cohorts. We therefore find evidence for a stable high-level cognitive signal rather than invariant feature-level explanations. This distinction matters because agreement in a dominant feature does not imply that the complete decision rule has transferred unchanged.

The imaging explanations showed a complementary pattern. Predictions and their associated attribution maps changed together, both when fusion was introduced within a cohort and when the same architecture was moved from ADNI to OASIS-3, rather than the explanation drifting independently of the prediction it accompanies. Explanation behavior, in other words, tracked the model's predictive regime rather than any fixed property of the architecture. This supports our central view of explainability as an empirical property to be evaluated under distribution shift, and it is consistent with our previous work showing that agreement between attribution methods can provide useful evidence about the robustness of explanations~\cite{brima2024saliency,singh2025unsupervised}.

\subsection{Limitations and future directions}

Our study has several limitations. First, the quantitative comparison of explanations is stronger for tabular features than for image attributions. We quantify agreement between SHAP and Integrated Gradients, but the visual analyses rely on representative cases and do not provide a cohort-level metric of spatial attribution similarity. Second, OASIS-3 differs from ADNI in diagnostic composition, recruitment, acquisition characteristics, and availability of clinical variables. These differences are necessary for testing external generalization but prevent us from attributing observed changes to a single source of distribution shift.

We also use clinical variables, including MMSE, that are closely related to the diagnostic process. Their strong contribution may therefore reflect information that is highly proximate to the target label rather than complementary disease biology. Future work should first evaluate a unified attribution framework based on Integrated Gradients across both tabular and imaging modalities, enabling more directly comparable attribution analyses across heterogeneous input types. Such an approach could also facilitate systematic assessment of attribution stability across model configurations and cohorts. In addition, gradient-based image explanations could be augmented with quantitative spatial analyses, such as localized attribution or blob-based characterization, to move beyond qualitative visualization toward cohort-level measurements of the magnitude, spatial distribution, and consistency of model attributions. Recent work has demonstrated such quantitative analysis of gradient-based explanations using anatomically informed points of interest and blob-level analysis of saliency maps~\cite{garret2025segmentation}. Applying and adapting such approaches to Alzheimer's disease classification could provide a more rigorous assessment of whether imaging explanations correspond to reproducible anatomical patterns.

Further work should quantify cross-cohort similarity of image attribution maps, evaluate explanation stability at the subject and cohort levels, and systematically ablate diagnosis-proximal cognitive variables. Additional independent cohorts would further allow us to determine whether the stable high-level cognitive signal and the cohort-dependent imaging explanations observed here persist across broader clinical settings. More broadly, future studies should investigate whether unified attribution methods and quantitative analysis of gradient-based explanations can improve the comparability, reproducibility, and clinical interpretability of multimodal explanations under distribution shift.

\section{Conclusion}
\label{sec:con}
We evaluated structural MRI, clinical variables, and their multimodal combinations for Alzheimer's disease diagnosis across ADNI and an independent OASIS-3 cohort. Clinical variables provided strong discrimination on ADNI, particularly for CN versus MCI, while cross-attention achieved the strongest ADNI performance for MCI versus AD. Under external evaluation, however, no fusion strategy consistently improved on the unimodal models. The imaging-only model achieved strong three-class discrimination on OASIS-3, whereas the relative performance of the clinical and multimodal models changed across diagnostic tasks.

Our explainability analyses similarly indicate that model behavior was dominated by cognitive measures in the tabular modality, while image-based attributions varied with the model and cohort. Taken together, these results do not support a general claim that multimodal fusion provides superior diagnostic performance over individual modalities. Rather, they show that the value of each modality depends on the diagnostic transition and data distribution. External validation and modality-specific explanation are therefore important when assessing whether multimodal Alzheimer's disease models learn information that is both predictive and transferable.

\backmatter

\section*{Declarations}
\subsection*{Funding}
This research was supported in part by the National Research Foundation of South Africa (Ref No. CSRP23040990793).
\subsection*{Availability of data and materials}
This study used two publicly available neuroimaging cohorts. The ADNI dataset is available at https://adni.loni.usc.edu/, and the OASIS-3 dataset is available at https://sites.wustl.edu/oasisbrains/home/oasis-3/. Access to both requires registration and approval in accordance with each repository's data use agreement.
\subsection*{Competing interests}
The authors declare that they have no competing interests

\subsection*{Ethics approval and consent to participate}
This study uses anonymized secondary data from the ADNI and OASIS3

\subsection*{Consent for publication}
Not applicable

\subsection*{Acknowledgments}
Not applicable

% \subsection*{Materials availability}

\subsection*{Code availability }

All code for data processing, model development, and analysis is publicly available at:
\url{https://github.com/yusufbrima/mxaiadni}

\subsection*{Author contributions}
Y.B. conceived the study, curated the data, developed the software, performed the formal analysis and investigation, and prepared the visualizations. Y.B. and M.A. developed the methodology and wrote the original manuscript draft. Y.B., M.A., L.R.N., and A.V. contributed to validation and reviewed and edited the manuscript. M.A. and A.V. supervised the study. Y.B. and M.A. managed the project administration, and M.A. secured funding. All authors reviewed and approved the final manuscript.



\begin{appendices}

\section{Data Characteristics}
\label{sec:appendix_data}
 
\begin{figure}[H]
    \centering
    \includegraphics[width=1.0\linewidth]{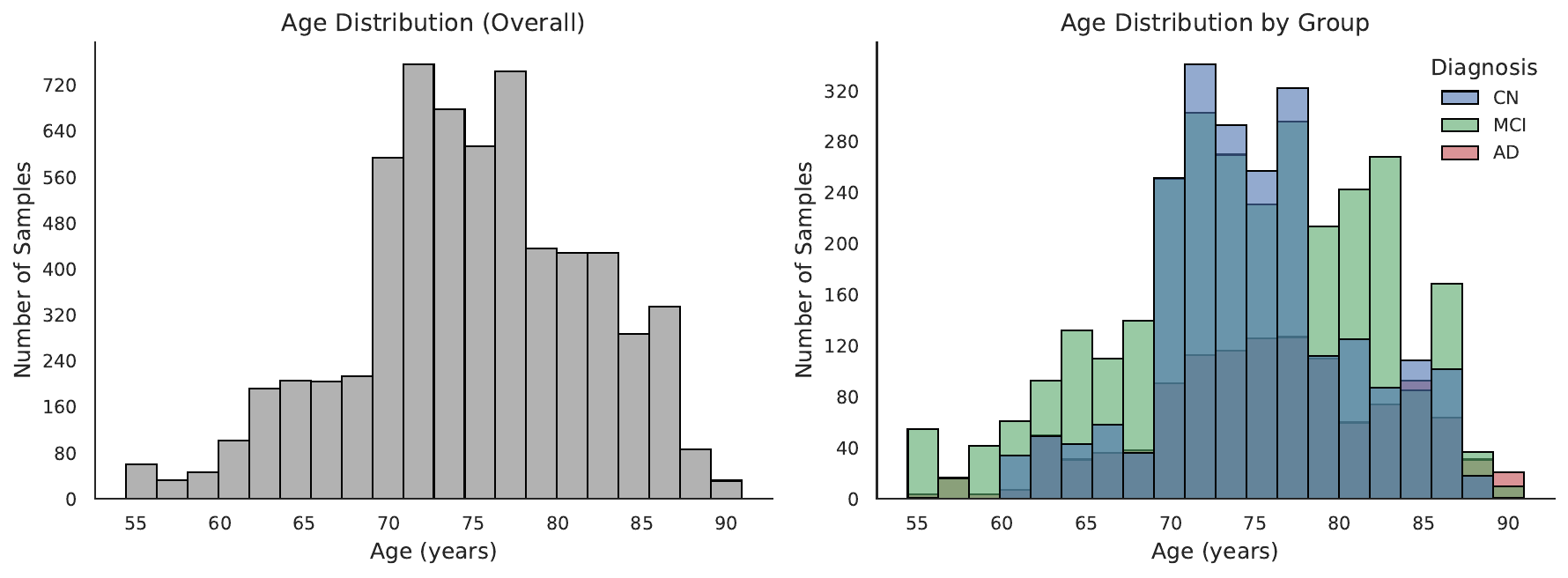}
    \caption{\textbf{Age Distribution.} Age distribution of ADNI subjects overall (left) and stratified by diagnostic group (right). The majority of subjects are aged 65--85 years, with broadly comparable age distributions across CN, MCI, and AD groups.}
    \label{fig:age_distribution}
\end{figure}
 
\begin{figure}[H]
    \centering
    \includegraphics[width=1.0\linewidth]{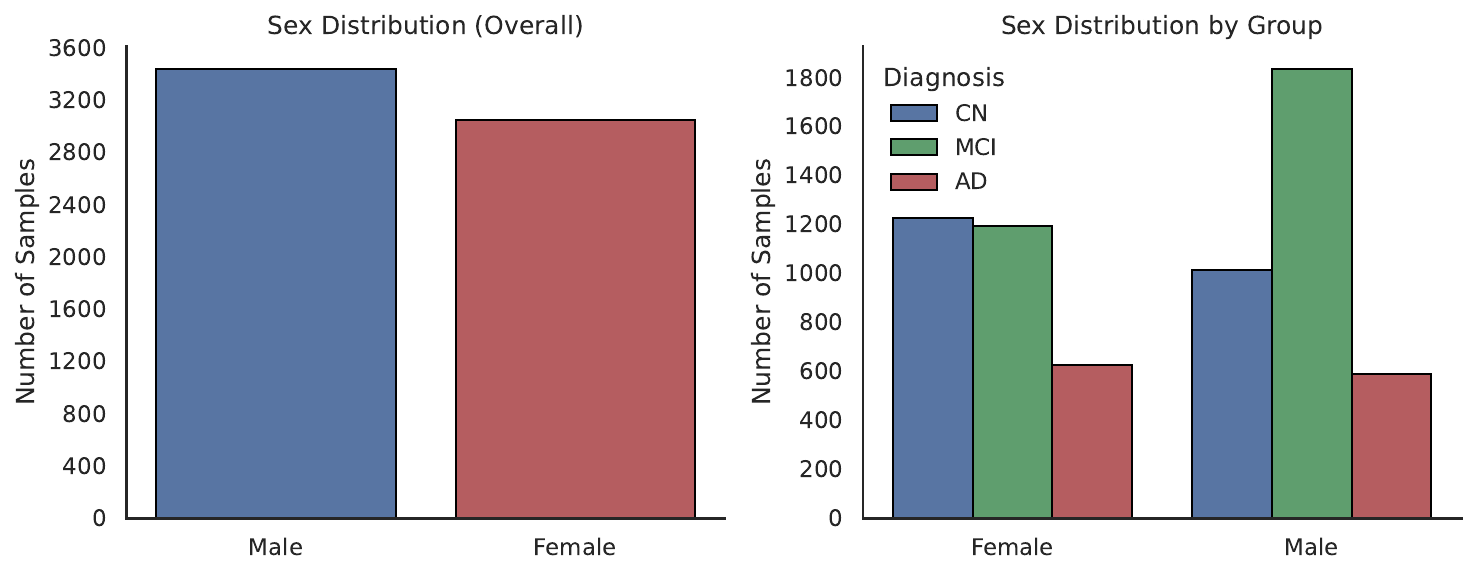}
    \caption{\textbf{Sex Distribution.} Sex distribution of ADNI subjects overall (left) and stratified by diagnostic group (right). The dataset is approximately balanced between female and male subjects, with MCI showing a slightly higher proportion of male subjects.}
    \label{fig:sex_distribution}
\end{figure}

\section{Tabular Variables}
\label{sec:appendix_tabular_vars}

% \begin{spacing}{1.15}
% \small
\begin{longtable}{@{}p{1.2in}p{1.3in}p{2.3in}@{}}
\caption{Harmonized Clinical, Demographical, and Neuropsychological Tabular Features Used for ADNI Training and OASIS-3 External Validation.}
\label{tab:tabular_features_appendix} \\
\toprule
\textbf{Feature Category} & \textbf{Feature Variable} & \textbf{Clinical Description} \\
\midrule
\endhead

\midrule
\multicolumn{3}{r}{\textit{Continued on next page}} \\
\bottomrule
\endfoot

\bottomrule
\endlastfoot

\textbf{Demographics \& Genetics} 
    & \texttt{AGE} & Patient age at baseline assessment. \\
    & \texttt{PTEDUCAT} & Years of formal education completed. \\
    & \texttt{Sex} & Biological sex of the participant (categorical predictor). \\
    & \texttt{APOE4} & Number of Apolipoprotein E $\varepsilon4$ alleles (0, 1, or 2). \\

\addlinespace
\textbf{Global Cognitive Assessment}
    & \texttt{MMSE\_bl} & Mini-Mental State Examination total score at baseline assessment. \\

\addlinespace
\textbf{Executive Function \& Processing Speed}
    & \texttt{DIGITSCOR\_bl} & Digit Symbol Substitution Test score at baseline, measuring processing speed and attention. \\
    & \texttt{TRABSCOR\_bl} & Trail Making Test Part B completion score at baseline, assessing executive function and cognitive flexibility. \\

\midrule
\textbf{Diagnostic Target}
    & \texttt{DX\_bl} & Clinical diagnostic classification used as prediction target: (\texttt{CN}), (\texttt{MCI}), or (\texttt{AD}). \\

\end{longtable}
% \end{spacing}

\section{Medical Image Pre-processing Pipeline}
\label{sec:appendix_mri_preprocess}

To reduce anatomical variability, intensity inhomogeneity, and computational overhead across heterogeneous acquisition sites and cohorts, all structural T1-weighted sMRI volumes were subjected to a standardized, deterministic pre-processing workflow. Individual volumes were first converted to 32-bit floating-point precision, followed by N4 bias field correction to mitigate smooth spatial intensity inhomogeneities associated with magnetic field and radio-frequency coil effects. Non-brain tissue was subsequently removed using an automated skull-stripping procedure to reduce the influence of non-diagnostic anatomical features on downstream visual representations and attribution maps. To reduce variability in global head orientation and anatomical positioning, the skull-stripped volumes were spatially registered to the standard MNI152 template space using affine registration, establishing a common stereotaxic coordinate system across subjects and cohorts. Following registration, voxel-wise Z-score intensity normalization was applied independently to each volume, standardizing the within-volume intensity distribution to approximately zero mean and unit variance. The same pre-processing procedure was applied to ADNI and OASIS-3 without cohort-specific intensity or contrast calibration; consequently, residual differences in scanner and acquisition-dependent contrast distributions between cohorts were retained. Finally, the normalized volumes were resampled to a fixed tensor shape of $128 \times 128 \times 128$ voxels, providing standardized inputs for the deep learning multimodal framework.

The pre-processing pipeline was implemented in Python 3.12.13 using SimpleITK 2.5.5, NiBabel 5.4.2, ANTsPyX 0.6.3, and ANTsPyNet 0.3.2. TensorFlow 2.21.0 and DeepBrain 0.1 were used for automated brain extraction. SimpleITK was used for intensity processing, N4 bias field correction, spatial registration, and volumetric resampling.

\section{Clinical Longitudinal Data Integration and Multimodal Alignment}
\label{sec:appendix_data_alignment}

To build a unified multimodal mapping framework, preprocessed structural 3D neuroimaging volumes were systematically linked with longitudinal EHR matrices derived from the cohort database. The processing pipeline dynamically parses file path metadata descriptors to extract unique participant identifiers and maps them against corresponding clinical registries. Demographic predictors and clinical classifications are standardized, including mapping biological sex descriptors and translating clinical dementia designations to a uniform indicator matching the target classification groups. Duplicate image files and missing clinical entries are strictly removed from the active data streams to preserve dataset integrity. To resolve temporal discrepancies inherent to heterogeneous real-world clinical data acquisition, a nearest-visit matching mechanism evaluates absolute time distances between individual image scan timestamps and clinical assessment logs. This synchronizes each 3D MRI scan with its nearest longitudinal clinical encounter to establish high temporal continuity. Finally, samples with missing target labels are omitted, and strict duplicate validation checks ensure that each unique volumetric tensor maps directly to a single, complete clinical observation, generating a consistent multimodal dataset optimized for robust representation learning.

\section{ADNI ScoreCAM Explainability}
\begin{figure}[H]
    \centering
    \includegraphics[width=1.0\linewidth]{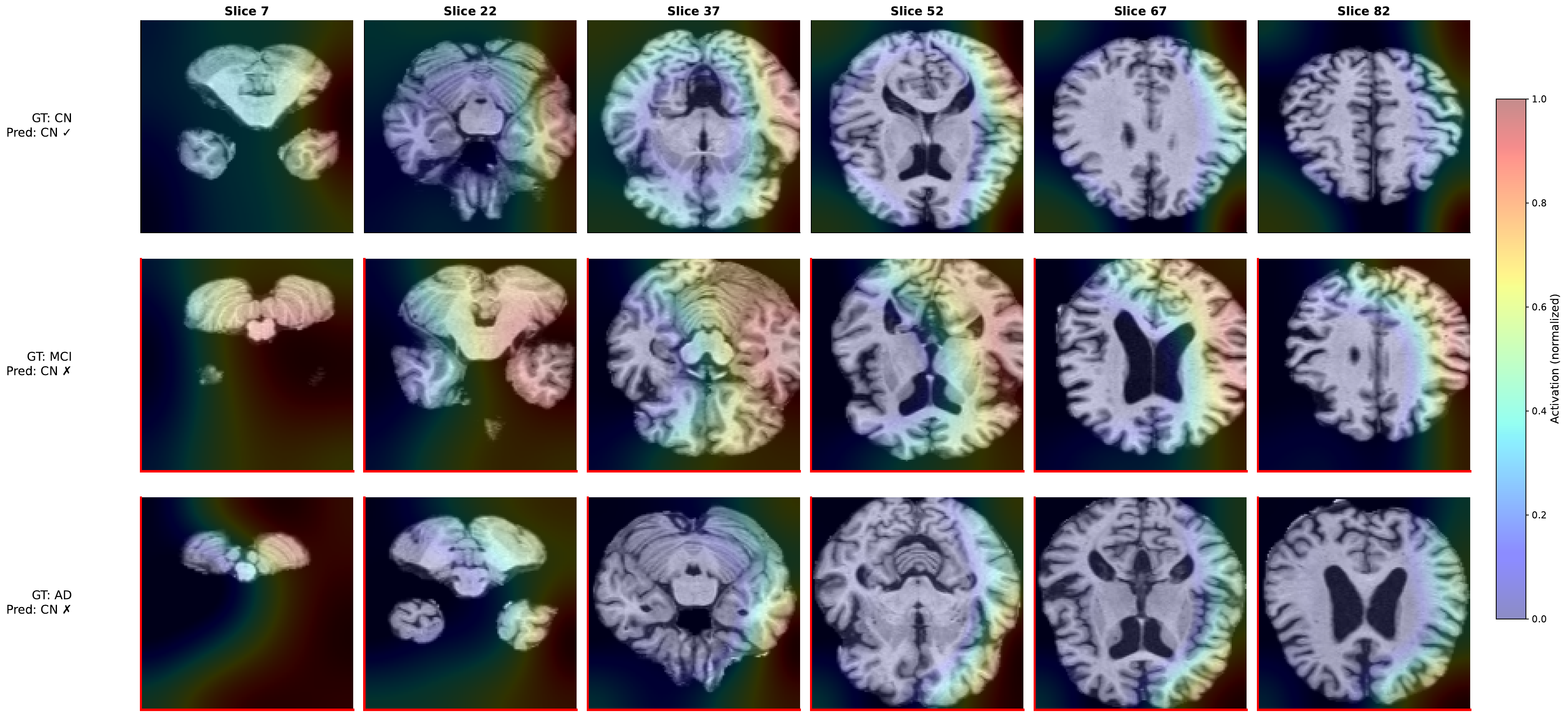}
    \caption{\textbf{ADNI Score-CAM Attributions.} Score-CAM attribution maps for the vision-only model on the same ADNI subjects shown in Fig.~\ref{fig:gradcampp_grid}. The CN subject is correctly classified, whereas the MCI and AD subjects are misclassified as CN.}
    \label{fig:scorecam_grid}
\end{figure}

\begin{figure}[H]
    \centering
    \includegraphics[width=1.0\linewidth]{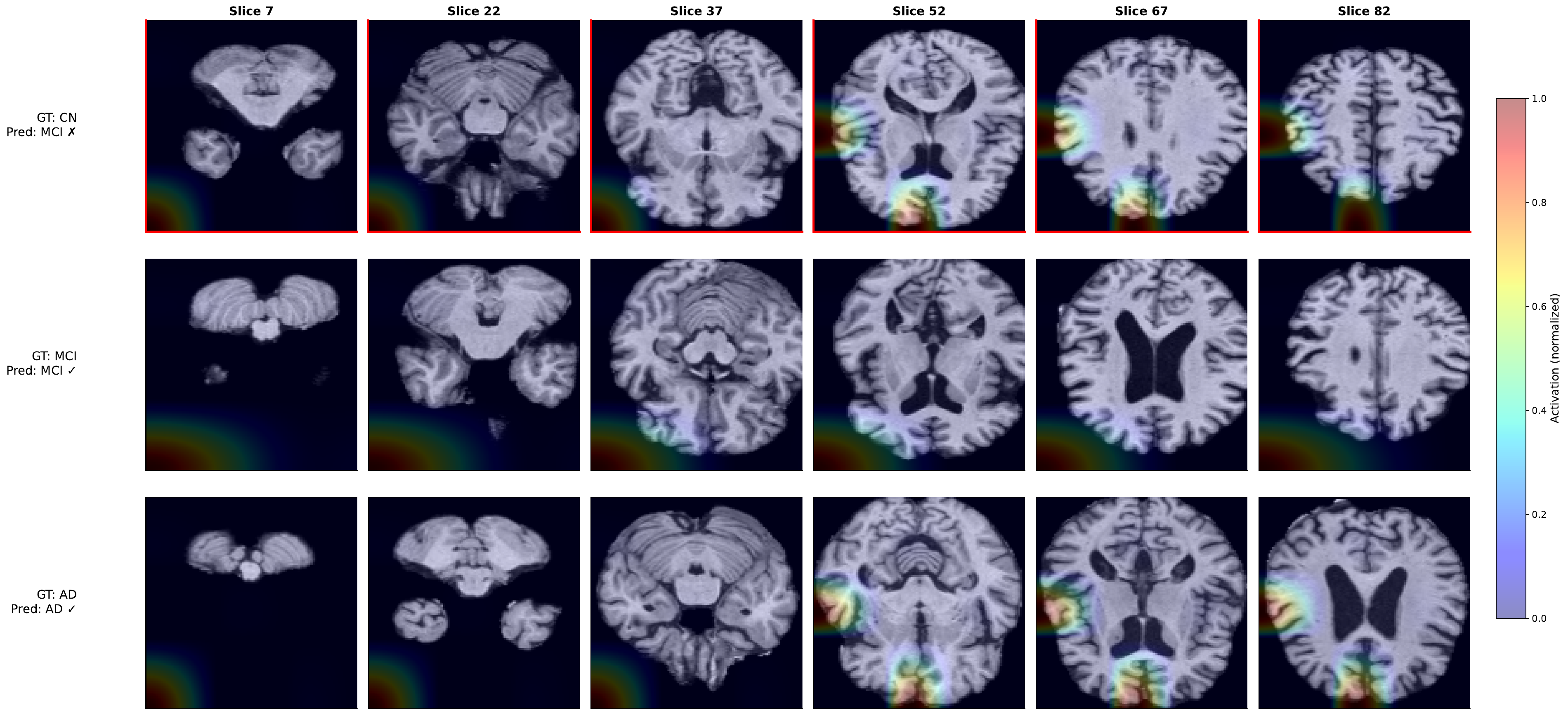}
    \caption{\textbf{ADNI Concatenation-fusion Score-CAM.} Score-CAM attribution maps for the cross-attention-fusion multimodal model on the same ADNI subjects.}
    \label{fig:scorecam_grid_multimodal_concat}
\end{figure}

\begin{figure}[H]
    \centering
    \includegraphics[width=1.0\linewidth]{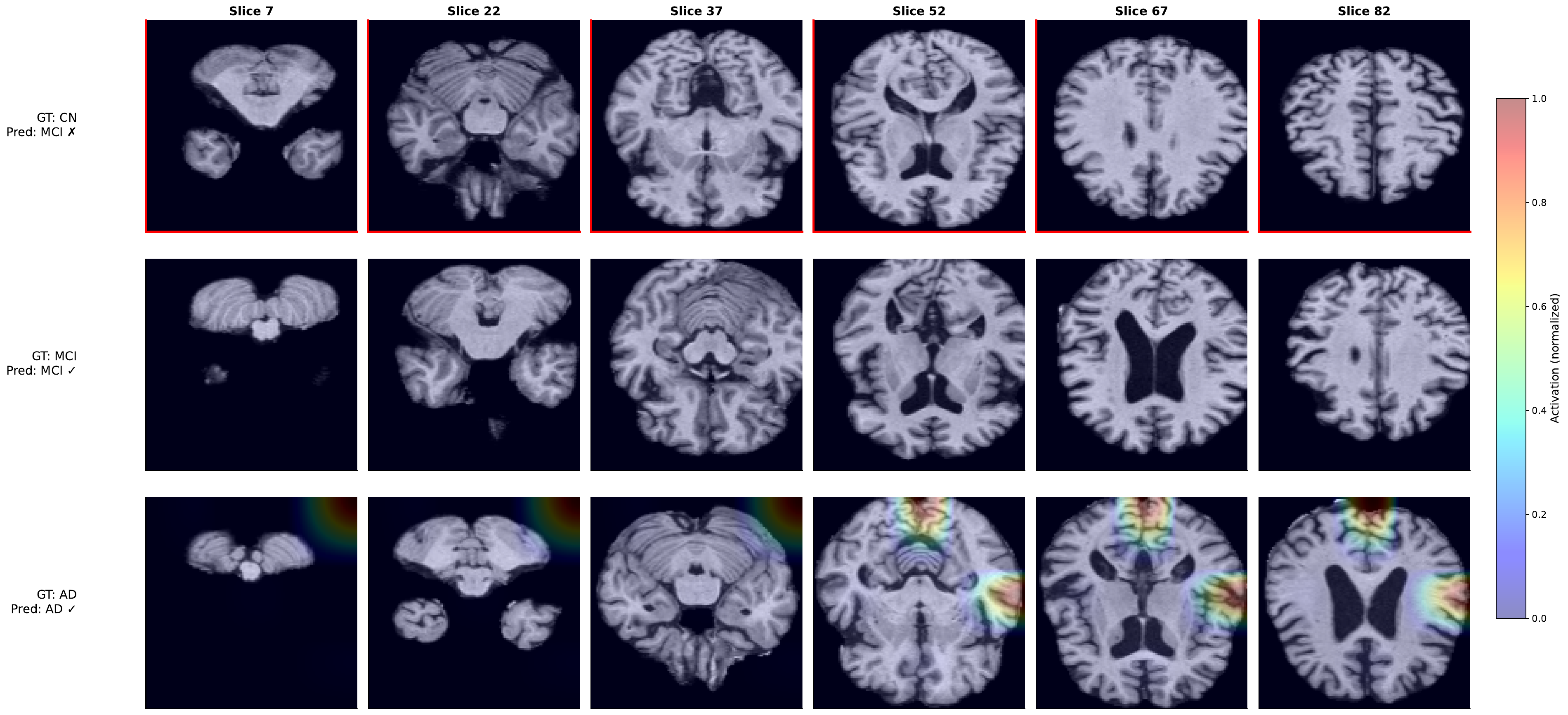}
    \caption{\textbf{ADNI Cross-Attention Score-CAM.} Score-CAM attribution maps for the cross-attention-fusion multimodal model on the same ADNI subjects.}
    \label{fig:scorecam_grid_multimodal_cross_attn}
\end{figure}

\section{OASIS3 Explainability}
\subsection{Tabular Explainability}
\begin{figure}[H]
    \centering
    \includegraphics[width=0.450\linewidth]{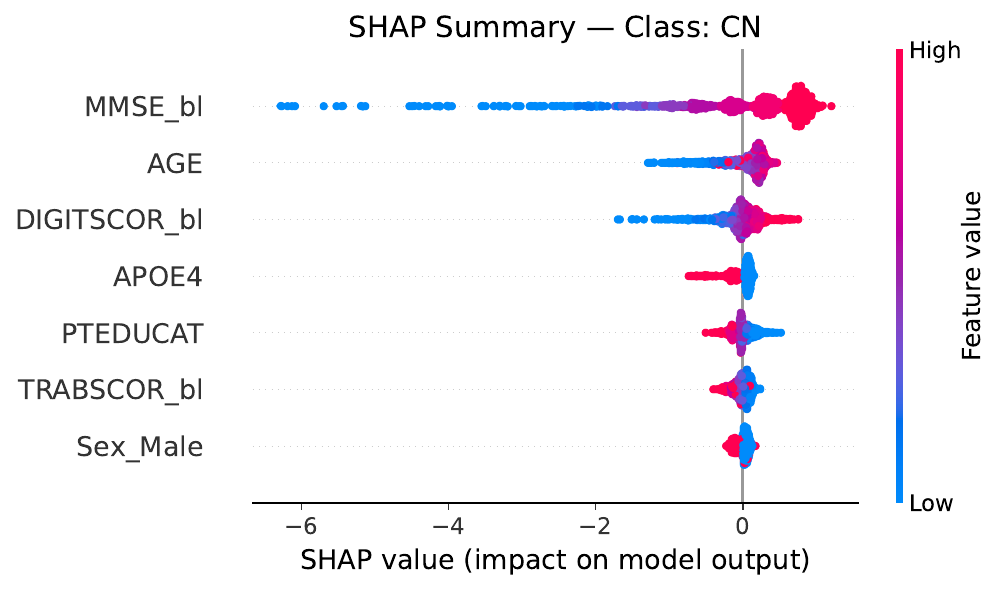}
    \includegraphics[width=0.450\linewidth]{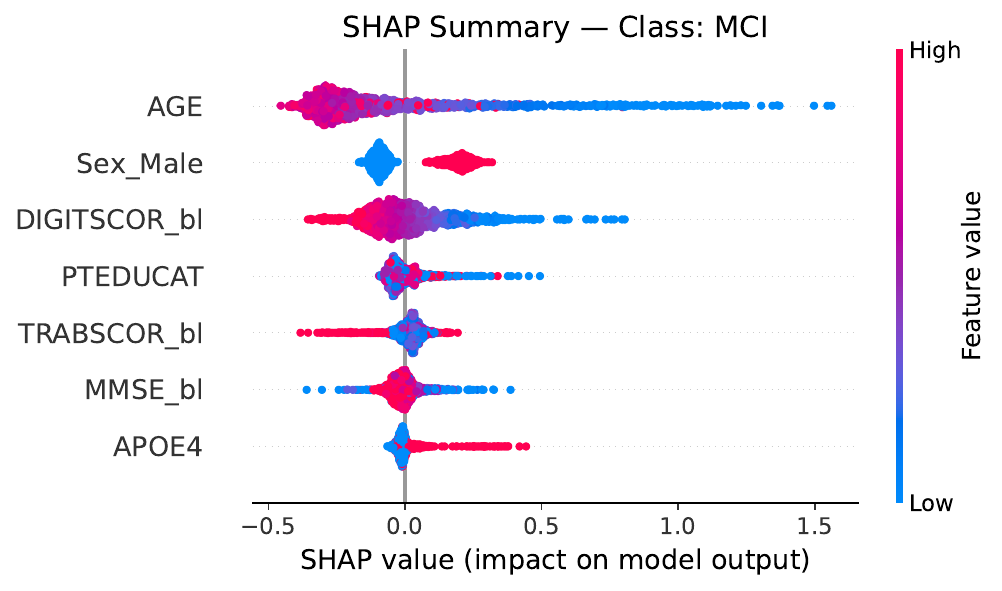}
    \includegraphics[width=0.450\linewidth]{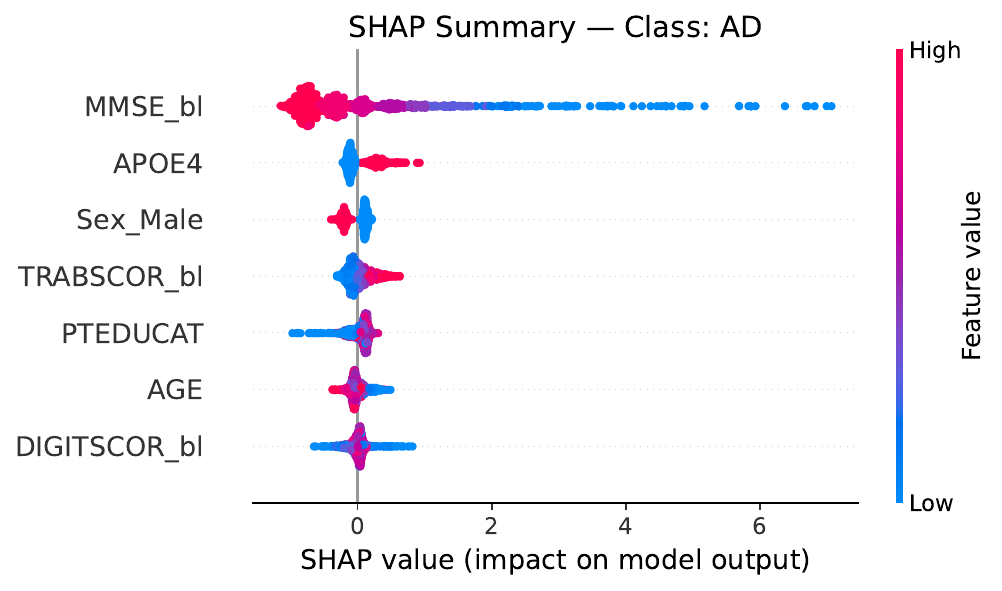}
    \caption{\textbf{OASIS-3 SHAP Summary.} SHAP summary plots for the individual diagnostic target classes (CN, MCI, and AD), outlining local feature attributions and impact profiles on model output distribution.}
    \label{fig:shap_summary_AD_oasis3}
\end{figure}

\begin{figure}[H]
    \centering
    \includegraphics[width=0.450\linewidth]{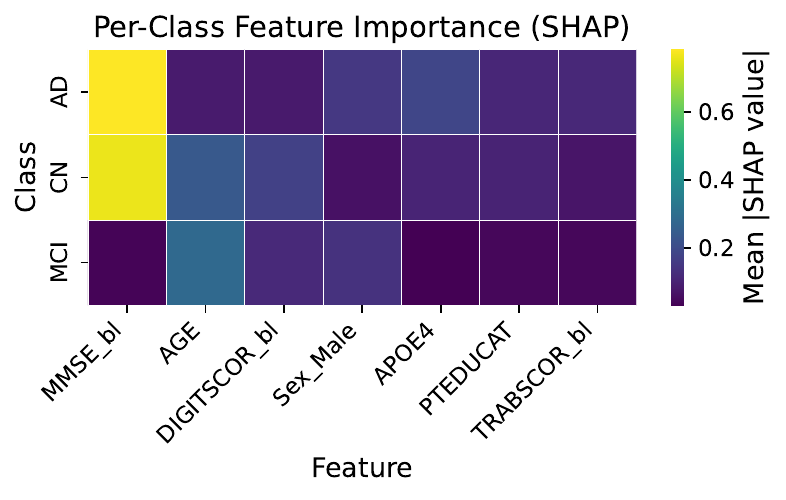}
    \includegraphics[width=0.450\linewidth]{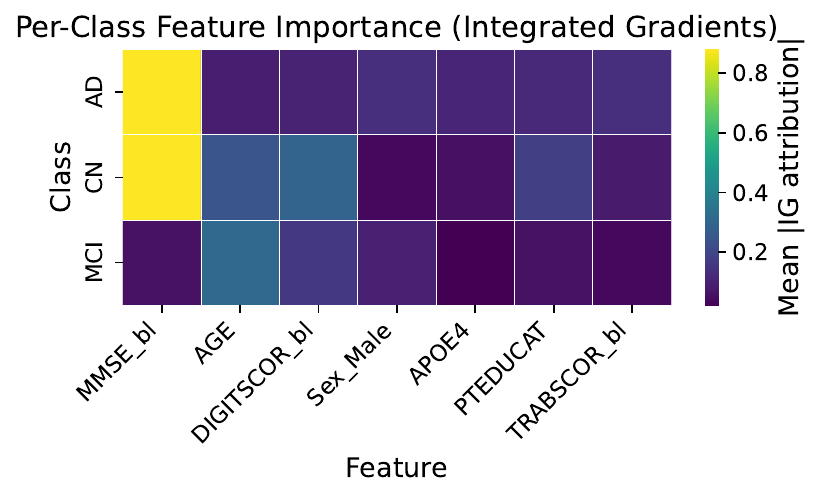}
    \caption{\textbf{OASIS-3 Global Feature Importance.} Per-class global feature importance from SHAP (left) and Integrated Gradients (right) in OASIS-3.}
    \label{fig:shap_importance_heatmap_by_class_oasis3}
\end{figure}

\subsection{Visual Explainability}
\begin{figure}[H]
    \centering
    \includegraphics[width=1.0\linewidth]{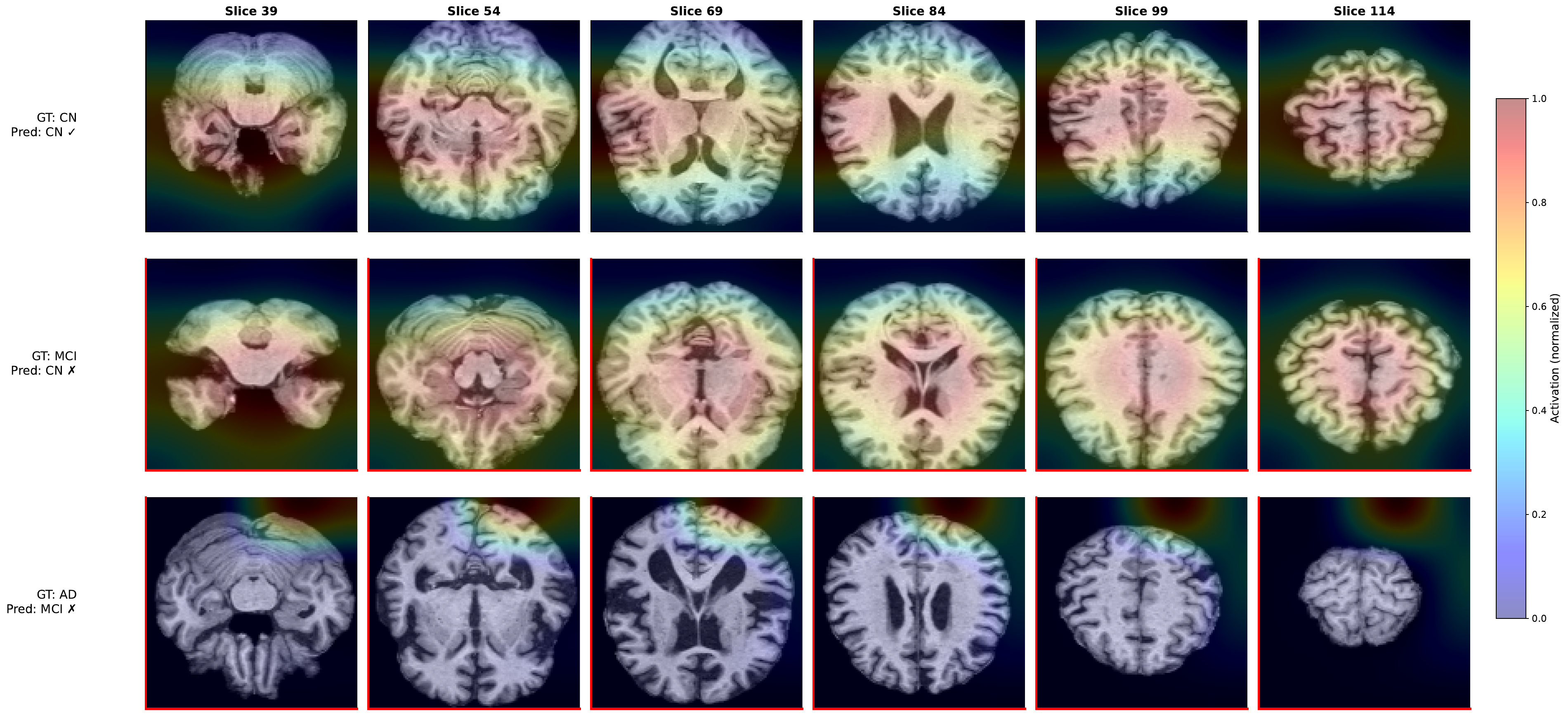}
    \caption{\textbf{OASIS-3 Vision-Only Grad-CAM++.} Grad-CAM++ attribution maps for the vision-only model on representative OASIS-3 subjects, overlaid on six axial slices per subject. The CN subject is correctly classified, while the MCI subject is misclassified as CN and the AD subject as MCI.}
    \label{fig:gradcampp_grid_oasis3}
\end{figure}

\subsection{Multimodal Explainability}
\begin{figure}[H]
    \centering
    \includegraphics[width=1.0\linewidth]{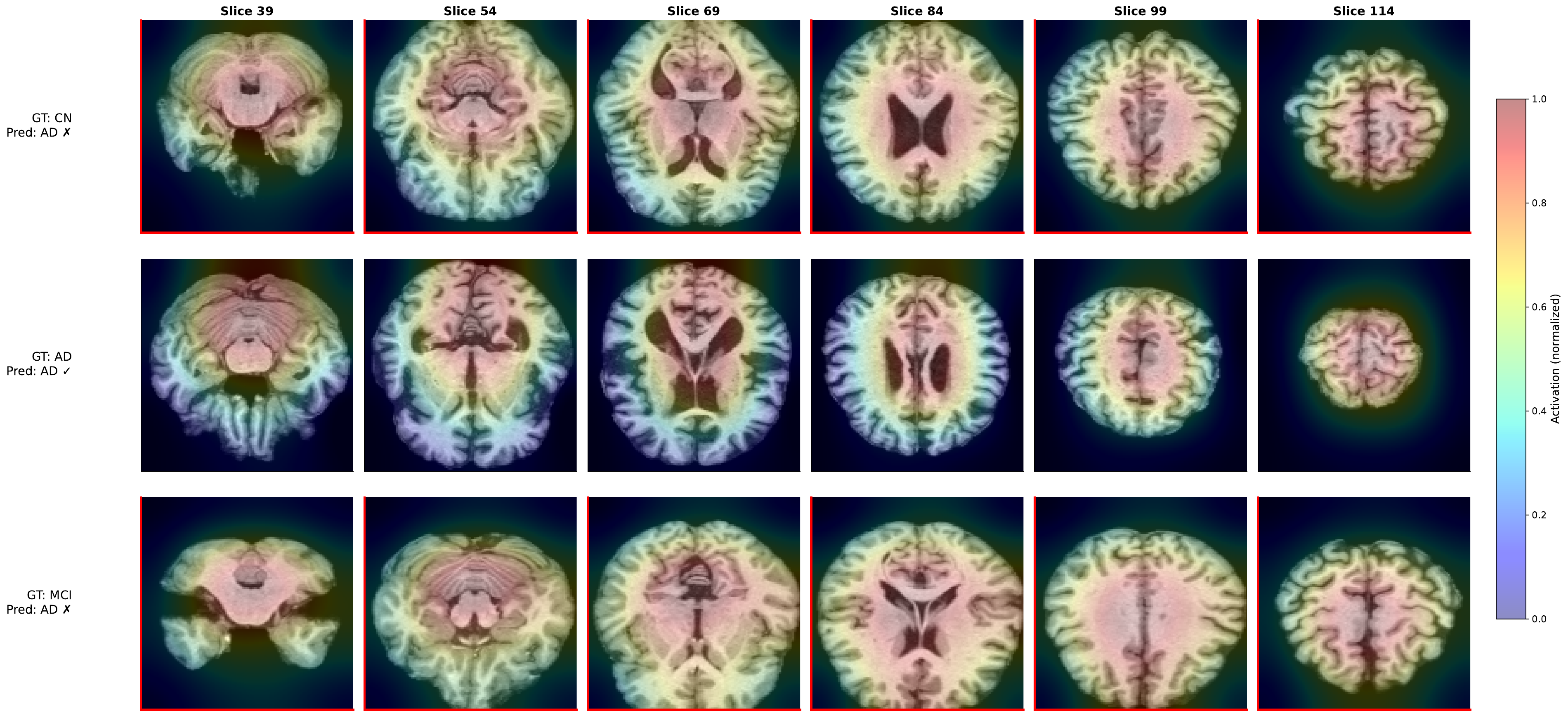}
    \caption{\textbf{OASIS-3 Concatenation-fusion Grad-CAM++.} Grad-CAM++ attribution maps for the Concatenation-fusion multimodal model on the same OASIS-3 subjects shown in Fig.~\ref{fig:gradcampp_grid_oasis3}. The CN and MCI subjects are misclassified as AD, while the AD subject is correctly classified.}
    \label{fig:gradcampp_grid_multimodal_concat_oasis3}
\end{figure}
\begin{figure}[H]
    \centering
    \includegraphics[width=1.0\linewidth,height=0.7\textheight, keepaspectratio]{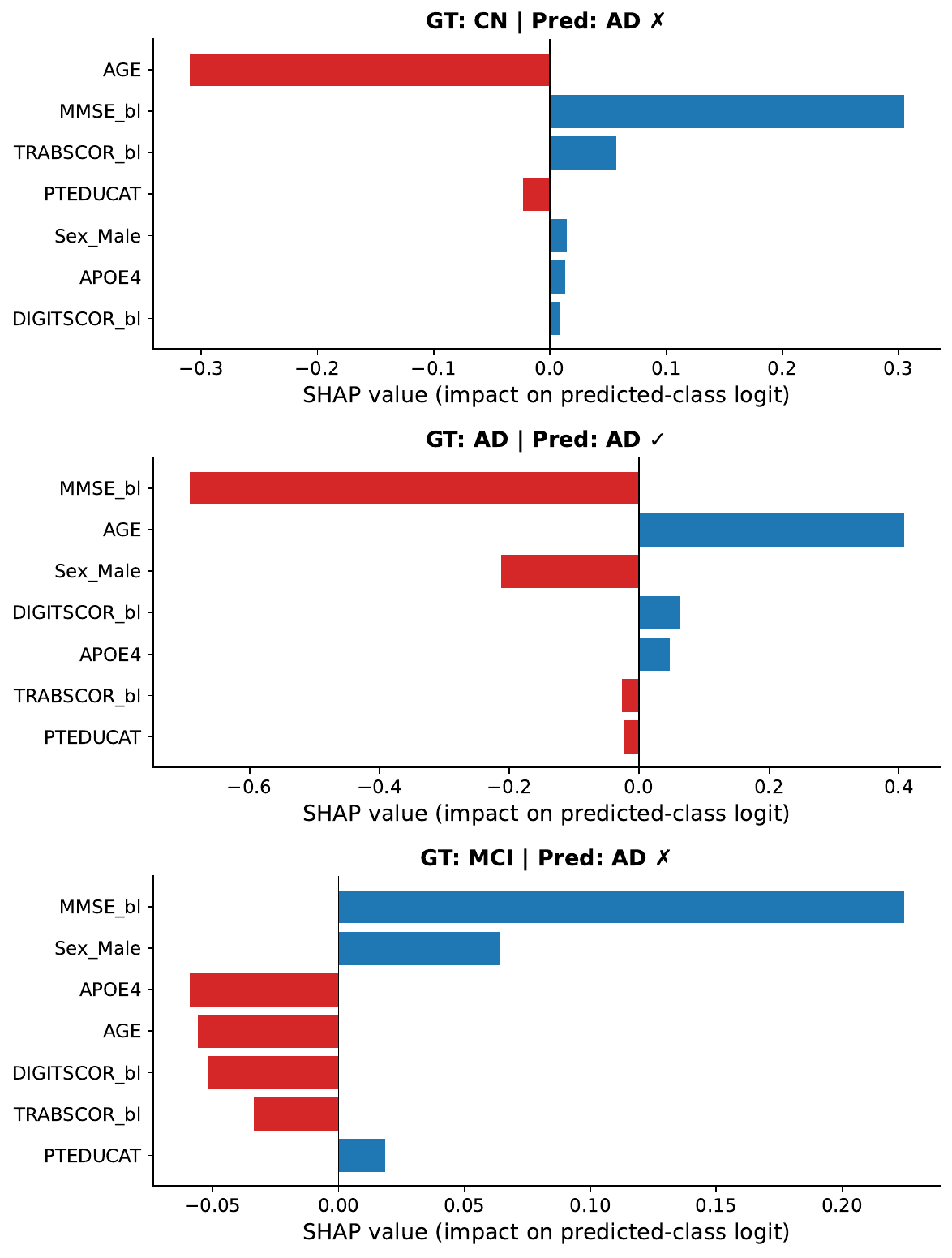}
    \caption{\textbf{OASIS-3 Concatenation-fusion SHAP.} SHAP feature attributions for the predicted-class of the tabular branch of the Concatenation-fusion model for the same OASIS-3 subjects shown in Fig.~\ref{fig:gradcampp_grid_multimodal_concat_oasis3}. Blue bars indicate positive and red bars negative contributions to the predicted-class.}
    \label{fig:tabular_shap_per_class_concat_oasis3}
\end{figure}

\begin{figure}[H]
    \centering
    \includegraphics[width=1.0\linewidth]{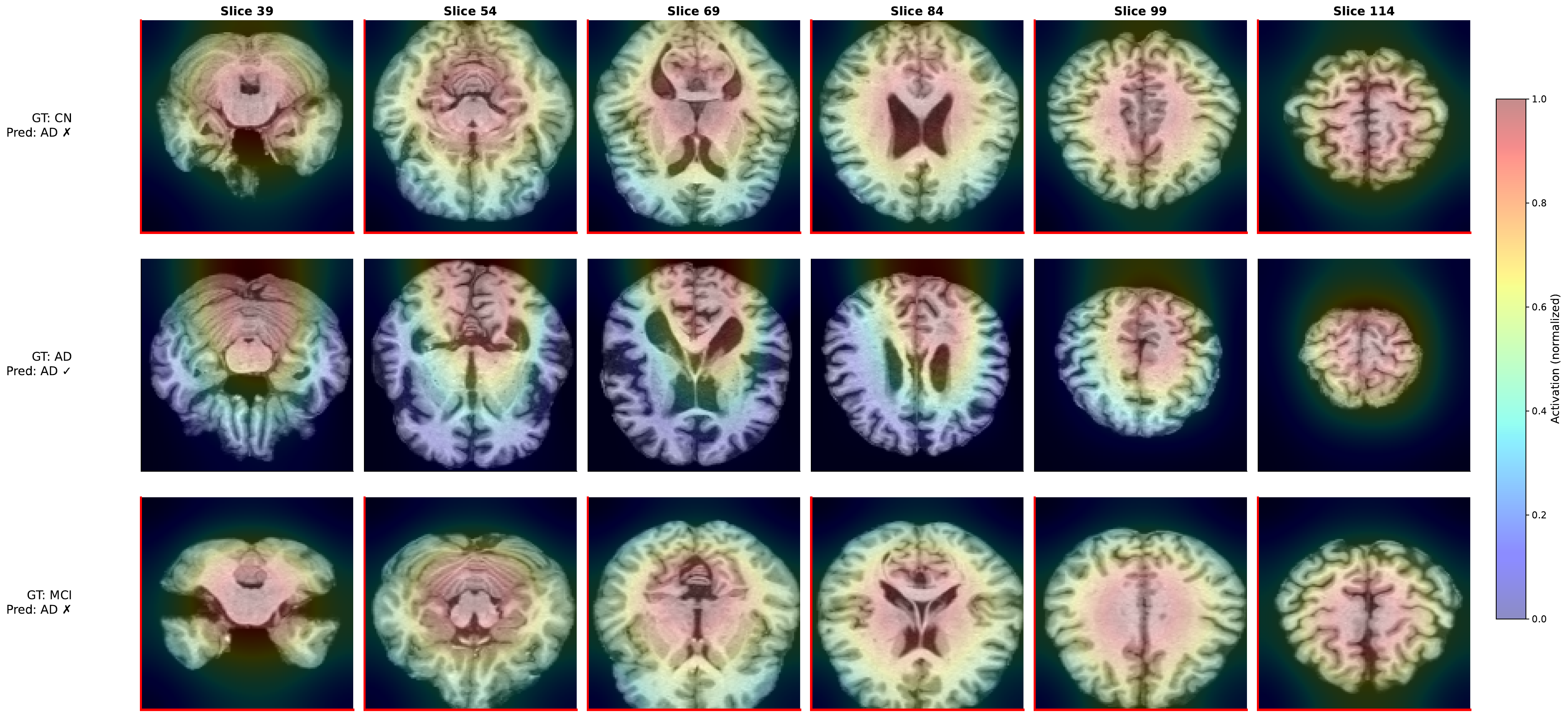}
    \caption{\textbf{OASIS-3 Cross-Attention Grad-CAM++.} Grad-CAM++ attribution maps for the cross-attention-fusion multimodal model on the same OASIS-3 subjects shown in Figs~\ref{fig:gradcampp_grid_oasis3} and~\ref{fig:gradcampp_grid_multimodal_concat_oasis3}. The CN and MCI subjects are misclassified as AD, while the AD subject is correctly classified.}
    \label{fig:gradcampp_grid_multimodal_cross_attn_oasis3}
\end{figure}

\begin{figure}[H]
    \centering
    \includegraphics[width=1.0\linewidth]{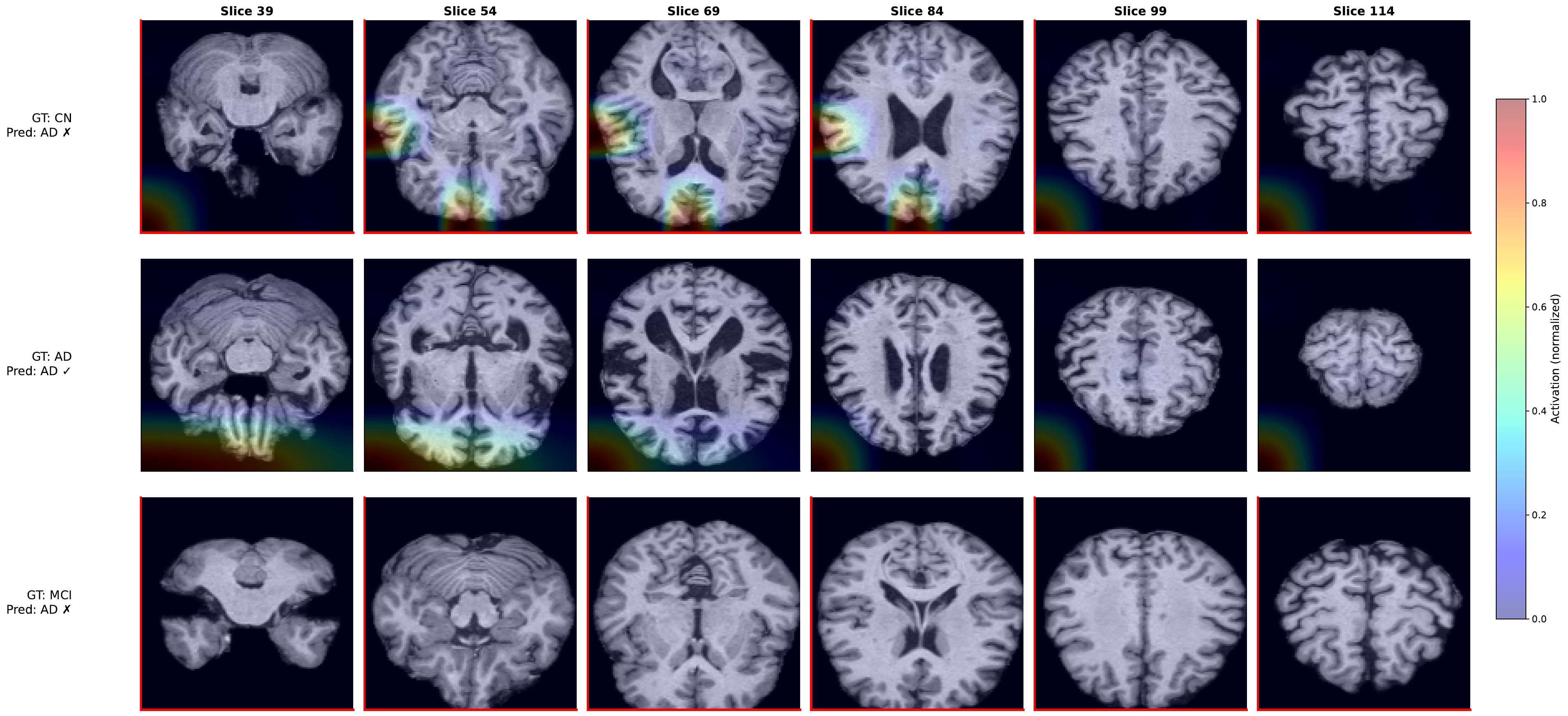}
    \caption{\textbf{OASIS-3 Concatenation-fusion Score-CAM.} Score-CAM attribution maps for the cross-attention-fusion multimodal model on the same OASIS-3 subjects.}
    \label{fig:scorecam_grid_multimodal_concat_oasis3}
\end{figure}

\begin{figure}[H]
    \centering
    \includegraphics[width=1.0\linewidth]{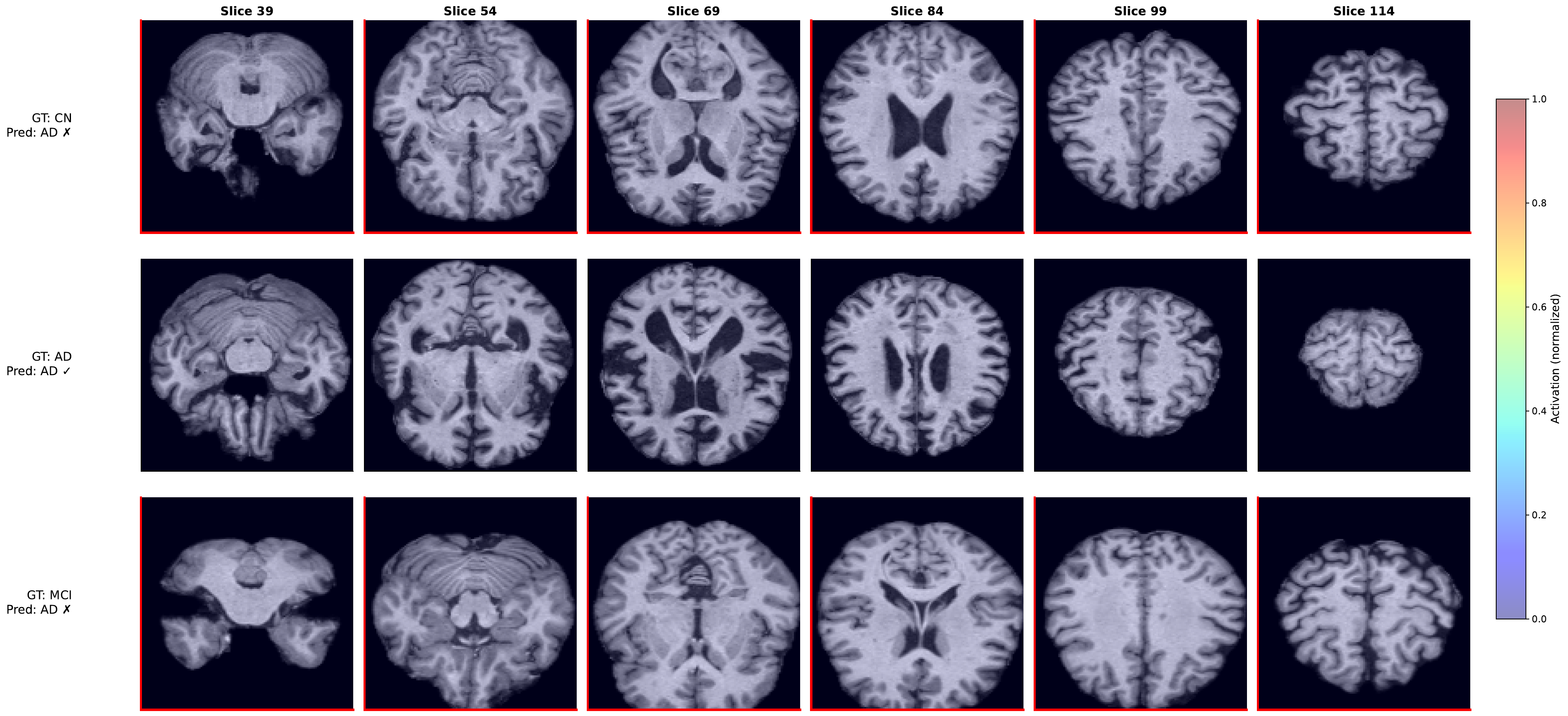}
    \caption{\textbf{OASIS-3 Cross-Attention Score-CAM.} Score-CAM attribution maps for the cross-attention-fusion multimodal model on the same OASIS-3 subjects.}
    \label{fig:scorecam_grid_multimodal_cross_attn_oasis3}
\end{figure}

\begin{figure}[H]
    \centering
    \includegraphics[width=1.0\linewidth]{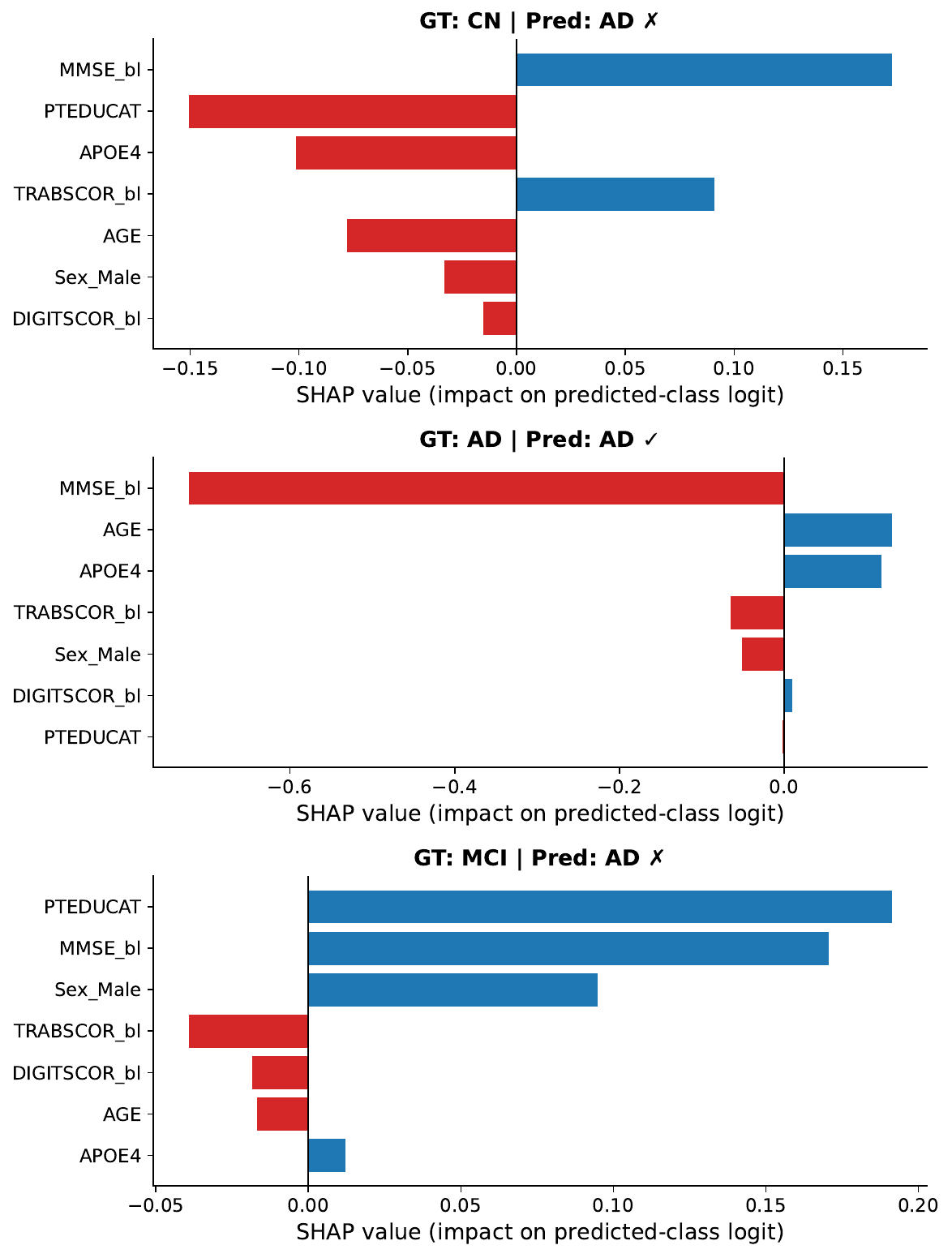}
    \caption{\textbf{OASIS-3 Cross-Attention SHAP.} SHAP feature attributions for the predicted-class of the tabular branch of the cross-attention-fusion model for the same OASIS-3 subjects shown in Fig.~\ref{fig:gradcampp_grid_multimodal_cross_attn_oasis3}. Blue bars indicate positive and red bars negative contributions to the predicted-class.}
    \label{fig:tabular_shap_per_class_cross_attn_oasis3}
\end{figure}

\end{appendices}

\bibliography{sn-bibliography}% common bib file

\end{document}